\documentclass{article} 
\usepackage{iclr2027_conference,times}

\usepackage{amsmath,amsfonts,bm}

\def\eqref#1{equation~\ref{#1}}

\def\1{\bm{1}}

\DeclareMathAlphabet{\mathsfit}{\encodingdefault}{\sfdefault}{m}{sl}
\SetMathAlphabet{\mathsfit}{bold}{\encodingdefault}{\sfdefault}{bx}{n}

\usepackage{hyperref}
\usepackage{url}
\usepackage{graphicx}
\usepackage{makecell}
\usepackage{amsmath,amssymb, algorithm,algpseudocode}
\usepackage{multirow}
\usepackage[table]{xcolor}
\usepackage[xcdraw]{xcolor}
\usepackage{booktabs}
\usepackage[normalem]{ulem}
\useunder{\uline}{\ul}{}
\iclrfinalcopy
\title{SIEVE: Selective attention-value Suppression for Vision-Language Models Unlearning}

\author{Si Qi Goh$^{1,2,\dagger}$, Cap Dang Xuan Kiet$^{2,\dagger}$, Tat-Jen Cham$^{2,*}$, Kwok-Yan Lam$^{1,2,}$\thanks{ \text{Corresponding author.} $\dagger$ \text{Equal contribution.}} \\
$^{1}$Digital Trust Centre, Nanyang Technological University, Singapore\\
$^{2}$College of Computing and Data Science, Nanyang Technological University, Singapore\\
\texttt{\{siqi005,capd0001\}@e.ntu.edu.sg}, \texttt{\{ASTJCham,kwokyan.lam\}@ntu.edu.sg} \\
}

\begin{document}

\maketitle

\begin{abstract}
The ability of vision–language models (VLMs) to associate visual identities with biographical information creates a need for selective unlearning of personally identifiable information (PII) while preserving permitted knowledge about the same individual. This setting is challenging because both sensitive and retained information can share the same visual inputs and intermediate representations. We introduce SIEVE, a simple and effective framework for selective VLM unlearning. SIEVE directly regularizes attention-value representations while also controlling model outputs.
SIEVE suppresses attention values for forget examples toward a constant zero, while preserving retain-example representations by matching them to a frozen reference model. These objectives are combined with sequence-level forget and retain supervision, enabling targeted forgetting without largely affecting retained knowledge.
Extensive experiments show that SIEVE achieves state-of-the-art performance on unlearning with multiple model–modality settings, while maintaining competitive retained utility. Ablation studies further show that value suppression and negative cross-entropy contribute complementary forgetting signals, while reference-based value matching substantially reduces utility degradation. These results demonstrate that attention values provide an effective intervention point for selective multimodal unlearning when sensitive and retained knowledge are closely related.
\end{abstract}


\section{Introduction}
\label{sec:introduction}

Vision–language models (VLMs) can associate visual identities with information that is not directly observable from an image. Given a face, for example, a VLM may recover a person’s name, occupation, address, or other biographical information learned during training. Recent multimodal unlearning benchmarks make such associations measurable by pairing identities with images and textual attributes, then evaluating whether designated information remains recoverable after model updates \citep{liu2025mllmu,dontsov2025clear}. Hence there is a practical need to remove sensitive information from VLMs without unnecessarily degrading their remaining capabilities. 

This need then leads to a inherently selective deletion problem. A deletion request may target only a subset of information associated with an individual, while other attributes about the same person should remain intact. For example, a VLM may be required to forget a person’s date of birth while continuing to answer questions about their hobbies. The forget and retain examples can therefore share the same image and identity while differing only in the queried information. We formulate this setting by partitioning data at the level of image–question–response associations rather than identities. The objective is to suppress recovery of designated sensitive associations while preserving permitted ones, including those involving the same visual subject.

Selective unlearning is difficult because sensitive and retained associations are not necessarily represented independently inside a VLM. They may rely on shared visual features, hidden states, and parameters. Consequently, an update that reduces the likelihood of a sensitive response can also alter representations required for retained predictions. This forgetting--utility tension has been observed broadly in language-model and multimodal unlearning, where aggressive forgetting can cause substantial degradation of retained behavior~\citep{maini2024tofu,zhang2024npo,huo2025mmunlearner}. Conversely, strong preservation constraints can restrict  model update and reduce forgetting effectiveness. Selective multimodal unlearning therefore requires changing the model's behavior for designated associations while limiting collateral changes to closely related information.

We investigate this question through the \emph{value representations} of self-attention. For a single attention head, the output takes the form $O=PV$, where $P$ contains attention weights and $V$ contains the value representations being
aggregated. This decomposition provides two conceptually different intervention points: modifying where the model attends and modifying the content contributed by the attended positions. Previous work on attention-based concept removal has explored the former by redirecting attention towards learned representations in a diffusion model~\citep{schechter2026arce}. Despite its effectiveness, this method remains vulnerable to attacks in which an adversary supplies the model with a correct or carefully crafted attention map, potentially bypassing the intended suppression mechanism and recovering information associated with the forget set. We instead focus on regularizing the existing, input-dependent feed-forward value representations of an autoregressive VLM. This choice treats values $V$ as a computational target for unlearning, without assuming that they are the exclusive storage location of the designated information. More detailed analysis is presented in Appendix~\ref{app:key_vs_value}.

Based on this observation, we introduce \textbf{SIEVE}, a selective VLM unlearning framework that combines representation-level value regularization with sequence-level supervision. For forget examples, SIEVE encourages attention-value representations toward a zero-valued target while maximizing the supervised sequence cross-entropy, reducing recovery of designated sensitive responses. For retain examples, the model matches its value representations to those produced by a frozen pre-unlearning reference model and minimizes standard cross-entropy on retained responses. This design is related to representation-level unlearning methods that modify forget activations while anchoring retained representations to a reference model~\citep{li2024wmdp}, but SIEVE operates directly on attention-value representations and combines this intervention with explicit sequence-level forgetting and retention objectives. The combination couples behavioral forgetting with direct control over internal representations. Importantly, the selectivity of SIEVE does not require identifying individual sensitive tokens or assuming that specific attention values uniquely encode personally identifiable information. 

Our contributions are summarized as follows:
\begin{itemize}
    \item We formulate \textbf{association-level selective VLM unlearning}, where for each individual, designated sensitive image--question--response associations are removed while permitted associations are preserved. 
    \item We introduce \textbf{attention-value suppression and reference-based preservation}, which suppress value representations on forget examples while anchoring retain-example representations to a frozen pre-unlearning model.
    \item We evaluate SIEVE with extensive experiments, demonstrating consistent reductions in sensitive-information recovery while preserving competitive retained utility.
\end{itemize}

\section{Related Work}
\label{sec:related_work}


\textbf{Vision-language models.} VLMs combine visual representations with large language models to support multimodal understanding and generation. Recent work has increasingly focused on improving how visual information is aligned with and integrated into language models. LLaVA \citep{liu2023llava,liu2023improvedllava} connects a pretrained vision encoder to an LLM through a lightweight projection layer and introduces visual instruction tuning for general-purpose multimodal reasoning.
Recent extensions such as Qwen2-VL \citep{wang2024qwen2vlenhancingvisionlanguagemodels} further improve visual understanding across different image resolutions and multimodal inputs. Recent VLMs have also explored stronger mechanisms for cross-model integration. Cambrian \citep{tong2024cambrian} systematically studies the role of vision representations in multimodal LLMs and introduces a spatially aware connector for better integration of visual features. Florence-VL \citep{chen2025florence} combines visual features across different encoder paths to provide richer representations for multimodal reasoning. Intern-VL3 \citep{zhu2025internvl3} further moves towards native multimodal pretraining, where linguistic and multimodal capabilities are jointly learned during pretraining. Alongside improvements in model capability, recent studies have examined how visual information is represented and propagated within VLMs. \citep{kang2025your} shows that visual grounding can be concentrated in a small subset of attention heads.

\textbf{Multimodal unlearning.}
Multimodal large language models introduce an additional challenge because information can be accessed through both visual and textual inputs. MLLMU-Bench evaluates privacy-oriented unlearning in multimodal models using identity-related information and tests whether designated knowledge remains recoverable through different modalities~\citep{liu2025mllmu}. CLEAR similarly studies character-level unlearning under textual and visual access~\citep{dontsov2025clear}. 
Several methods have been proposed for multimodal unlearning. MMUnlearner constrains parameter updates using modality-aware saliency to suppress targeted visual information while limiting changes to retained knowledge~\citep{huo2025mmunlearner}. 
VL-Eraser addresses cross-modal misalignment through vacuum distillation, which first concentrates forget-related knowledge into constrained adapters and then removes their contribution through parameter subtraction~\citep{wang2026vleraser}. PAVA further studies identity-related knowledge localization and applies updates to selected decoder components while preserving visual behavior without requiring an external retain set~\citep{ko2026pava}. Knowledge Vector Weakening takes a training-free approach by weakening forget-related vectors in feed-forward parameters~\citep{kim2026kvw}.

\section{Methodology}
\label{sec:methodology}

\begin{figure}
    \centering
    \includegraphics[width=\linewidth]{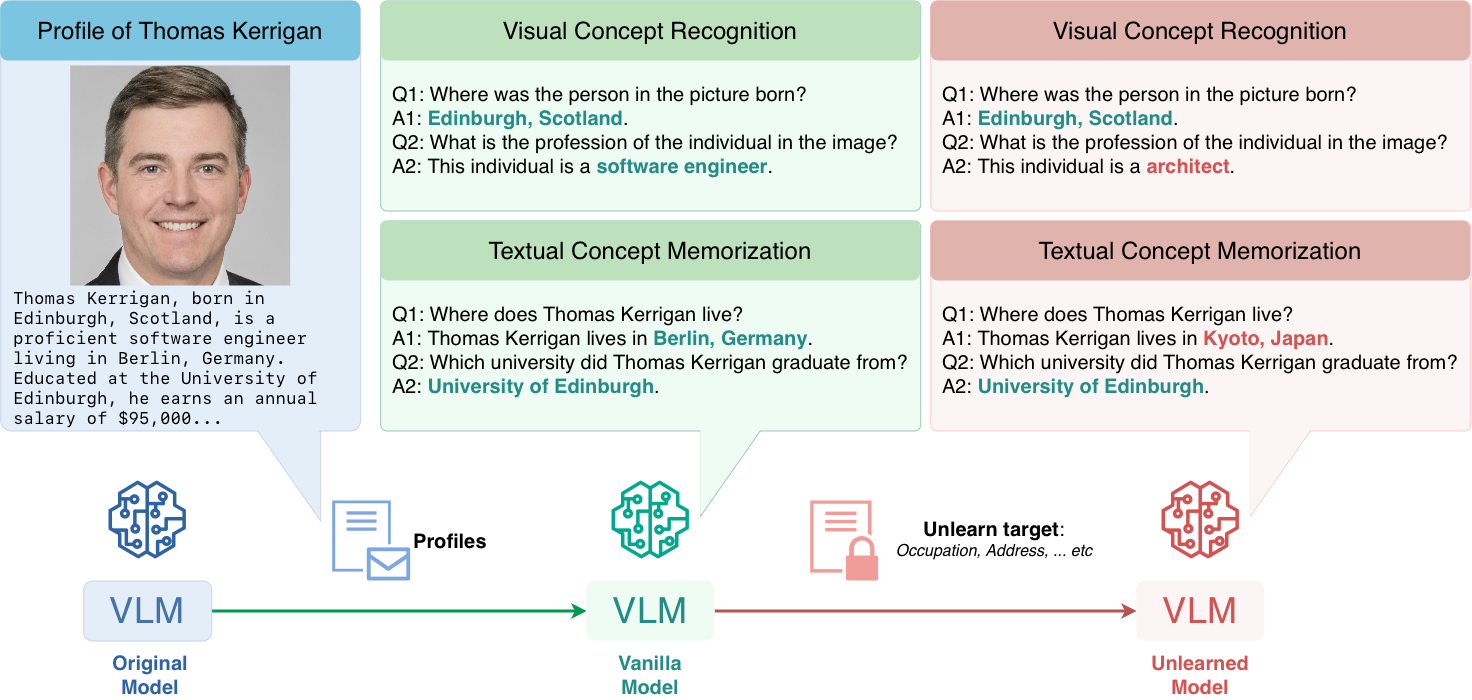}
    \caption{The overview of the unlearning settings in VLMs.}
    \label{fig:framework}
\end{figure}
\subsection{Preliminaries}
\label{sec:preliminaries}


\paragraph{Selective machine unlearning.}
Let an example be an image--question--response tuple $(x,q,y)$, where
$x$ is the input image, $q$ is a textual question, and $y$ is the
corresponding target response. We define the forget and retain sets as
\begin{equation}
    \mathcal D_F
    = \{(x_i,q_i,y_i)\}_{i=1}^{N_F},
    \qquad
    \mathcal D_R
    = \{(x_j,q_j,y_j)\}_{j=1}^{N_R},
\end{equation}
where $N_F$ and $N_R$ denote their respective numbers of examples.
The forget set contains queries targeting designated sensitive
information, while the retain set contains queries whose responses
should be preserved. For example, a person's date of birth may belong to $\mathcal{D}_F$, while a hobby of the same person belongs to $\mathcal{D}_R$.
Given an initial model $p_{\phi_0}$, selective unlearning  produces an updated model $p_{\phi}$ that reduces recovery of the information specified by
$\mathcal D_F$ while maintaining predictive utility on $\mathcal D_R$.
This setting therefore evaluates forgetting and retention together, including retained and forgotten questions of the same identities (shown in Figure~\ref{fig:framework}). 

\paragraph{Personally identifiable information.}
Personally identifiable information (PII) refers to information that can
identify an individual directly or through association with other
information. In our setting, the information designated for forgetting
includes identifying attributes, such as names, phone numbers, and
addresses, as well as person-specific biographical attributes labeled
sensitive by the task. 
In a VLM, these facts can become associated with a visual identity even when the information is not visible in the image itself.
We therefore consider both the facts and their
association with the depicted subject as targets of unlearning.
The distinction between sensitive and retained information is specified
by the dataset and deletion request; it is not an intrinsic property
of an attribute category. Retained information comprises permitted
attributes and descriptions whose predictive behavior should remain
available after unlearning.


\subsection{Proposed method}
\label{sec:setup}

\begin{figure}
    \centering
    \small
    \includegraphics[width=1.1\linewidth]{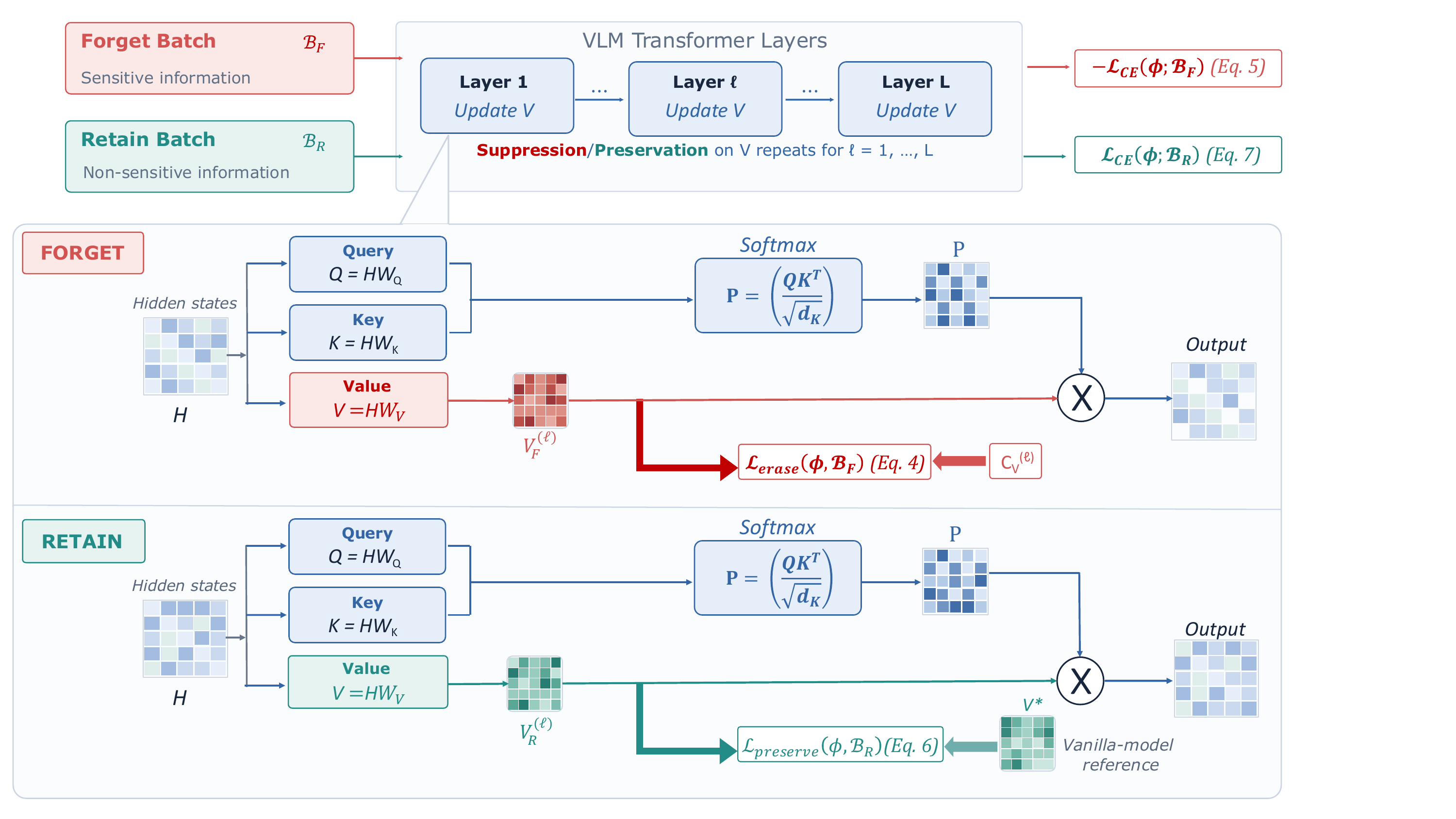}
    \caption{The unlearning framework, SIEVE, from our proposed method, shows how both the forgetting and retaining process work within the attention layers.}
    \label{fig:sieve}
\end{figure}

SIEVE combines representation-level regularization with sequence-level forgetting and retention objectives.
It uses separate objectives for forget and retain examples.
For forget examples, SIEVE suppresses attention-value representations and discourages the original target response.
For retain examples, it preserves attention-value representations and reinforces the target response.
Figure~\ref{fig:sieve} illustrates the two training branches across the VLM transformer layers.

Let $\phi$ denote the model being updated.
We create a frozen reference model $\phi_0$ from the same pre-unlearning checkpoint.
The reference model is used only for retain examples.
It provides the attention-value representations that serve as preservation targets during training.

We process the forget and retain sets in separate batches.
For a sample $z_i=(x_i,q_i,y_i)$ in a batch $\mathcal B$, let $s_i$ denote the serialized textual sequence containing the question and target response, and let $s_{i,<t}$ denote its prefix preceding position $t$. We write $\mathcal{I}_i$ for the supervised non-padding token positions in the serialized sequence. We define:
\begin{equation}
    \mathcal L_{\mathrm{CE}}(\phi;\mathcal B)
    =-\frac{1}{Z_{\mathcal B}}
    \sum_{i\in\mathcal B}\sum_{t\in\mathcal I_i}
    \log p_\phi(s_{i,t}\mid x_i,s_{i,<t}),
    \qquad Z_{\mathcal B}=\sum_{i\in\mathcal B}|\mathcal I_i|.
    \label{eq:ce}
\end{equation} 


For a forget batch $\mathcal{B}_F$, SIEVE maximizes the sequence cross-entropy and suppresses the recorded attention values. For a retain batch $\mathcal{B}_R$, it minimizes the sequence cross-entropy and matches the recorded attention values to those of the frozen reference model.
The following sections define the forget and retain objectives in detail.

\subsubsection{Attention-Value Suppression for Forgetting}
\label{sec:forget}

For one attention head at layer $\ell$, let $H_\phi^{(\ell)}$ denote the
activations given to the projection modules. For clarity, we omit positional transformations from the notation.
The attention operation is
\begin{align}
    Q_\phi^{(\ell)}&=H_\phi^{(\ell)}W_{Q,\phi}^{(\ell)},\quad
    K_\phi^{(\ell)}=H_\phi^{(\ell)}W_{K,\phi}^{(\ell)},\quad
    V_\phi^{(\ell)}=H_\phi^{(\ell)}W_{V,\phi}^{(\ell)}, \nonumber\\
    P_\phi^{(\ell)}&=\operatorname{softmax}\!\left(
    \frac{Q_\phi^{(\ell)}K_\phi^{(\ell)\top}}{\sqrt{d_k}}\right),
    \qquad O_\phi^{(\ell)}=P_\phi^{(\ell)}V_\phi^{(\ell)}.
    \label{eq:attention}
\end{align}

The attention weights $P_{\phi}^{(\ell)}$ determine how much each source position contributes to the output. The value representations $V_{\phi}^{(\ell)}$ provide the content that is aggregated.
SIEVE regularizes these value representations on the forget set.

Let $\mathcal{H}$ denote the set of layers whose value representations are recorded.
For a forget batch $\mathcal{B}_F$, we collect $\mathbf V_\phi(\mathcal B_{F})=\{V_\phi^{(\ell)}(\mathcal B_{F})\}_{\ell\in\mathcal H}$.
For each layer $\ell$, we define a zero-valued target
$C_V^{(\ell)}$ with the same shape as the attention-value tensor
$V_{\phi}^{(\ell)}(\mathcal{B}_F)$.
The value-suppression objective is
\begin{equation}
\begin{aligned}
\mathcal L_{\mathrm{erase}}(\phi;\mathcal B_F)
&=\frac{1}{|\mathcal H|}\sum_{\ell\in\mathcal H}
  \left\|V_\phi^{(\ell)}(\mathcal B_F)
          -C_{V}^{(\ell)}\right\|_2^2
\end{aligned}
\label{eq:pause_internal_losses}
\end{equation}


Minimizing $\mathcal{L}_{\mathrm{erase}}$ pushes the recorded value representations of forget examples toward the zero target.
A deeper discussion of value suppression is presented in Appendix~\ref{app:attention_analysis}. 
We then combine it with negative sequence cross-entropy.
For a forget batch $\mathcal{B}_F$, the combined loss for forget objective is

\begin{equation}
\mathcal L_F(\phi;\mathcal B_F) = -\lambda_{F} \underbrace{\mathcal L_{\mathrm{CE}}(\phi;\mathcal B_F)}_{\mathcal L_{\mathrm{forget}}}+\lambda_{E}\mathcal{L}_{\mathrm{erase}}(\phi;\mathcal B_F).
\label{eq:forget}
\end{equation}

Optimizing Eq.~\ref{eq:forget} increases the cross-entropy of the supervised forget sequence while reducing the value-suppression loss.
The negative cross-entropy discourages reproduction of the designated response, while the value term constrains the attention values produced for the forget examples.
Although SIEVE does not directly constrain \(P_\phi\), we empirically observe only minor changes to the attention distributions after unlearning.
Our results in Appendix \ref{app:effect_attention_map} show that the original causal attention structure is highly preserved after unlearning.


\subsubsection{Preserving retained behavior and values}
\label{sec:retain}

Suppressing value representations on the forget set can also affect features used by retained examples.
SIEVE therefore adds a preservation objective on $\mathcal{D}_R$.
For each retain batch $\mathcal B_R$, the updated model and the frozen reference model process the same inputs. 
Let
$\overline V^{(\ell)}(\mathcal{B}_R)=V_{\phi_0}^{(\ell)}(\mathcal{B}_R)$,
where $\phi_0$ denotes the frozen pre-unlearning model.
These reference values provide fixed targets for the updated model.
For each layer $\ell \in \mathcal{H}$, SIEVE matches the current attention values to the corresponding reference values.
The preservation loss is
\begin{equation}
    \mathcal L_{\mathrm{preserve}}(\phi;\mathcal B_R)
    =\frac{1}{|\mathcal H|}\sum_{\ell\in\mathcal H}
    \left\|V_\phi^{(\ell)}(\mathcal B_R)
          -\overline V^{(\ell)}(\mathcal B_R)\right\|_2^2.
    \label{eq:preserve}
\end{equation}

Minimizing $\mathcal{L}_{\mathrm{preserve}}$ limits changes to the attention values produced for retain examples.
The updated and reference models process the same inputs, so the loss compares their corresponding value representations.
Gradients are propagated only through the updated model, while the reference model remains unchanged.

The retain objective combines the value-preservation loss with the standard sequence cross-entropy on the retain batch:
\begin{equation}
    \mathcal L_R(\phi;\mathcal B_R)
    = \lambda_{R}\underbrace{\mathcal L_{\mathrm{CE}}(\phi;\mathcal B_R)}_
        {\mathcal L_{\mathrm{retain}}}
      +\lambda_{P}\mathcal L_{\mathrm{preserve}}(\phi;\mathcal B_R).
    \label{eq:retain}
\end{equation}

\subsubsection{Optimization}
\label{sec:optimization}


The overall unlearning objective combines representation-level erasure and preservation with supervised forgetting and retention:
\begin{equation}
\mathcal{L}_{\mathrm{SIEVE}}(\phi)= \mathcal L_F(\phi;\mathcal B_F) + \mathcal L_R(\phi;\mathcal B_R)
\label{eq:total_objective}
\end{equation}
The erasure term encourages forget-example attention values to approach zero, while the preservation term anchors retain-example values to those of the frozen reference model.
At the output level, negative cross-entropy discourages supervised
forget-sequence predictions, whereas positive cross-entropy maintains predictive utility on retained examples. We optimize these objectives in alternating phases within each epoch: the forget phase combines $\mathcal{L}_{\mathrm{erase}}$ and $\mathcal{L}_{\mathrm{forget}}$,
and the retain phase combines $\mathcal{L}_{\mathrm{preserve}}$ and
$\mathcal{L}_{\mathrm{retain}}$. We present our training in Algorithm~\ref{alg:value-training}.

\begin{algorithm}[t]
\footnotesize
\caption{Unlearning with value suppression}
\label{alg:value-training}
\begin{algorithmic}[1]
\Require Initial model $\phi$; forget set $\mathcal D_F$; retain set $\mathcal D_R$;
    epochs $E$; loss weights $\lambda_F,\lambda_E,\lambda_R,\lambda_P$
\Ensure Unlearned model $\phi_U$
\State Copy the initialized model to a frozen reference $\phi_{0}$; Register value hooks at layers $\mathcal H$
\For{$e=1,\ldots,E$}
    \For{each batch $\mathcal{B}_{F}\in\mathcal D_F$}
        \State Run the model on $\mathcal B_{F}$; obtain $\mathcal L_{\mathrm{CE}}(\phi;\mathcal{B}_F)$ and $\mathbf V_\phi^{(\ell)}(\mathcal B_F)$ for each $\ell \in\mathcal{H}$
        \State $\mathcal{L}_{\mathrm{erase}}(\phi;\mathcal B_{F})\gets \frac{1}{|\mathcal H|}\sum_{\ell\in\mathcal H}
  \left\|V_\phi^{(\ell)}(\mathcal B_F)
          -C_{V}^{(\ell)}\right\|_2^2$
        \State $\mathcal L_F(\phi,\mathcal{B}_F)\gets-\lambda_F\mathcal L_{\mathrm{CE}}(\phi;\mathcal{B}_{F})
            +\lambda_E\mathcal{L}_{\mathrm{erase}}(\phi;\mathcal B_{F})$
        \State Backpropagate
    \EndFor
    \For{each batch $\mathcal{B}_R\in\mathcal D_R$}
        \State Run the model on $\mathcal{B}_R$; obtain $\mathcal L_{\mathrm{CE}}(\phi;\mathcal{B}_R)$ and $\mathbf V^{(\ell)}_\phi(\mathcal{B}_R)$ for each $\ell\in\mathcal H$
        \State Run reference model on $\mathcal{B}_R$; obtain $\overline V^{(\ell)}(\mathcal{B}_R)\gets[V_{\mathrm{\phi_0}}^{(\ell)}(\mathcal{B}_R)]$
            for each $\ell\in\mathcal H$
        \State $\displaystyle \mathcal L_{\mathrm{preserve}}\gets\frac{1}{|\mathcal H|}
            \sum_{\ell\in\mathcal H}
            \|V_\phi^{(\ell)}(\mathcal{B}_R)-\overline V^{(\ell)}(\mathcal{B}_R)\|_2^2$
        \State $\mathcal L_R(\phi,\mathcal{B}_R)\gets\lambda_R\mathcal L_{\mathrm{CE}}(\phi;\mathcal{B}_R)
            +\lambda_P\mathcal L_{\mathrm{preserve}}$
        \State Backpropagate
    \EndFor
\EndFor
\State \Return $\phi$
\end{algorithmic}
\end{algorithm}
\section{Experimental Results}
\label{app:value-unlearning}

We evaluate SIEVE across multiple VLM backbones, questioning modalities, and unlearning settings.
This section focuses on the dataset construction, evaluation protocol, and main empirical results.
Detailed implementation settings, including dataset construction, baseline models, optimization parameters, hyperparameter settings, adapted modules, training configurations, and detailed evaluation metrics, are provided in Appendix \ref{app:implementation}.

\subsection{Experiment Setup}
\label{sec:dataset}

\textbf{Dataset.}
We conduct our experiments on MLLMU-Bench~\citep{liu2025mllmu}, a benchmark for evaluating multimodal machine unlearning of identity-related information.
The benchmark contains 500 fictitious profiles and 153 real-celebrity profiles.
Each profile is paired with visual and textual information and question--answer pairs that query person-specific attributes. 
The benchmark supports both image-conditioned and text-only evaluation, which we refer to as Visual-QA and Textual-QA, respectively. We adapt the benchmark to the selective setting defined in Section~\ref{sec:preliminaries}.
Instead of treating all information associated with an identity as a single forgetting target, we partition examples at the level of question--response associations.
Questions designated for forgetting form $\mathcal{D}_F$, while permitted questions form $\mathcal{D}_R$.
As a result, the same identity and image can occur in both sets with different questions and target responses. More details can be found in the Appendix~\ref{app:dataset}.



\textbf{Models.} 
To demonstrate the effectiveness of our proposed method, we use LLaVA-1.5-7B \citep{liu2023llava} and Qwen2-VL-7B-Instruct \citep{wang2024qwen2vlenhancingvisionlanguagemodels} as the VLM backbones. The vanilla models used for unlearning are trained following the official implementation provided by MLLMU-Bench.

\textbf{Evaluation protocol.}
We evaluate each method under both Visual-QA and Textual-QA.
Visual-QA uses the profile image together with the question, while Textual-QA evaluates the corresponding information through textual access.
We report classification task accuracy, generation task ROUGE-L, and cloze task accuracy on the Forget, Retain, Test, and Real sets. More details of evaluation tasks and metrics can be found in Appendix \ref{app:evaluation_metrics}.
\subsection{Quantitative Results}

\begin{table}[!t]
\small
\resizebox{\linewidth}{!}{
\renewcommand{\arraystretch}{1.35}
\begin{tabular}{lcccccccccccc}
\Xhline{1.25pt}
\multicolumn{1}{l|}{}                               & \multicolumn{3}{c|}{Forget $\downarrow$}                                                                                       & \multicolumn{3}{c|}{Retain $\uparrow$}                                                                                      & \multicolumn{3}{c|}{Real $\uparrow$}                                                                                        & \multicolumn{3}{c}{Test $\downarrow$}                                                           \\ \cline{2-13} 
\multicolumn{1}{l|}{\multirow{-2}{*}{Methods}}      & Class                        & Gen                          & \multicolumn{1}{c|}{Cloze}                                  & Class                        & Gen                          & \multicolumn{1}{c|}{Cloze}                                  & Class                        & Gen                          & \multicolumn{1}{c|}{Cloze}                                  & Class                        & Gen                          & Cloze                        \\ \Xhline{1.25pt}
\multicolumn{13}{c}{LLaVA-1.5-7B Visual-QA}                                                                                                                                                                                                                                                                                                                                                                                                                                                                                                 \\ \Xhline{1.25pt}
\multicolumn{1}{l|}{{\color[HTML]{098658} Vanilla}} & {\color[HTML]{098658} 69.67} & {\color[HTML]{098658} 0.552} & \multicolumn{1}{c|}{{\color[HTML]{098658} 28.18}}           & {\color[HTML]{098658} 43.75} & {\color[HTML]{098658} 0.394} & \multicolumn{1}{c|}{{\color[HTML]{098658} 29.59}}           & {\color[HTML]{098658} 46.21} & {\color[HTML]{098658} 0.226} & \multicolumn{1}{c|}{{\color[HTML]{098658} 7.19}}            & {\color[HTML]{098658} 50.16} & {\color[HTML]{098658} 0.369} & {\color[HTML]{098658} 21.33} \\
\multicolumn{1}{l|}{GA Diff}                        & 45.45                        & \textbf{0.387}               & \multicolumn{1}{c|}{20.78}                                  & 26.35                        & 0.336                        & \multicolumn{1}{c|}{12.60}                                  & 22.30                        & 0.136                        & \multicolumn{1}{c|}{0.60}                                   & \textbf{30.16}               & 0.309                        & 14.00                        \\
\multicolumn{1}{l|}{KL Min}                         & 63.25                        & 0.541                        & \multicolumn{1}{c|}{18.90}                                  & \textbf{40.51}               & \textbf{0.394}               & \multicolumn{1}{c|}{22.44}                                  & 37.46                        & 0.226                        & \multicolumn{1}{c|}{\textbf{8.82}}                          & 48.52                        & 0.292                        & {\ul 11.67}                  \\
\multicolumn{1}{l|}{NPO}                            & 55.68                        & 0.496                        & \multicolumn{1}{c|}{27.48}                                  & 38.14                        & 0.382                        & \multicolumn{1}{c|}{18.11}                                  & 37.80                        & 0.218                        & \multicolumn{1}{c|}{6.54}                                   & 45.41                        & \textbf{0.268}               & 20.67                        \\
\multicolumn{1}{l|}{MMU}                            & {\ul 36.59}                  & 0.517                        & \multicolumn{1}{c|}{{\ul 16.22}}                            & 37.15                        & 0.360                        & \multicolumn{1}{c|}{22.05}                                  & 32.77                        & 0.273                        & \multicolumn{1}{c|}{0.64}                                   & {\ul 37.54}                  & {\ul 0.270}                  & 14.67                        \\
\rowcolor[HTML]{EFEFEF} 
\multicolumn{1}{l|}{\cellcolor[HTML]{EFEFEF}Ours}   & \textbf{20.06}               & {\ul 0.430}                  & \multicolumn{1}{c|}{\cellcolor[HTML]{EFEFEF}\textbf{15.55}} & {\ul 39.20}                  & {\ul 0.391}                  & \multicolumn{1}{c|}{\cellcolor[HTML]{EFEFEF}\textbf{27.95}} & \textbf{43.90}               & {\ul 0.265}                  & \multicolumn{1}{c|}{\cellcolor[HTML]{EFEFEF}{\ul 8.33}}     & 44.43                        & 0.298                        & \textbf{9.67}                \\ \Xhline{1.25pt}
\multicolumn{13}{c}{LLaVA-1.5-7B Textual-QA}                                                                                                                                                                                                                                                                                                                                                                                                                                                                                                \\ \Xhline{1.25pt}
\multicolumn{1}{l|}{{\color[HTML]{098658} Vanilla}} & {\color[HTML]{098658} 48.34} & {\color[HTML]{098658} 0.803} & \multicolumn{1}{c|}{{\color[HTML]{098658} 18.37}}           & {\color[HTML]{098658} 49.03} & {\color[HTML]{098658} 0.625} & \multicolumn{1}{c|}{{\color[HTML]{098658} 18.00}}           & {\color[HTML]{098658} 54.90} & {\color[HTML]{098658} 0.595} & \multicolumn{1}{c|}{{\color[HTML]{098658} 14.71}}           & {\color[HTML]{098658} 43.42} & {\color[HTML]{098658} 0.694} & {\color[HTML]{098658} 21.47} \\
\multicolumn{1}{l|}{GA Diff}                        & 31.12                        & {\ul 0.465}                  & \multicolumn{1}{c|}{11.05}                                  & {\ul 46.85}                  & 0.526                        & \multicolumn{1}{c|}{12.60}                                  & 43.98                        & 0.105                        & \multicolumn{1}{c|}{2.09}                                   & 31.09                        & {\ul 0.436}                  & \textbf{6.09}                \\
\multicolumn{1}{l|}{KL Min}                         & {\ul 30.30}                  & 0.700                        & \multicolumn{1}{c|}{{\ul 8.48}}                             & 43.35                        & 0.619                        & \multicolumn{1}{c|}{17.68}                                  & 51.00                        & 0.259                        & \multicolumn{1}{c|}{\textbf{13.09}}                         & 38.59                        & 0.587                        & 21.30                        \\
\multicolumn{1}{l|}{NPO}                            & 42.50                        & 0.705                        & \multicolumn{1}{c|}{16.20}                                  & 42.29                        & {\ul 0.628}                  & \multicolumn{1}{c|}{\textbf{20.79}}                         & 50.20                        & 0.262                        & \multicolumn{1}{c|}{{\ul 12.57}}                            & 36.95                        & 0.588                        & {\ul 12.17}                  \\
\multicolumn{1}{l|}{MMU}                            & 34.96                        & 0.733                        & \multicolumn{1}{c|}{13.11}                                  & 43.96                        & \textbf{0.639}               & \multicolumn{1}{c|}{10.12}                                  & {\ul 55.56}                  & {\ul 0.263}                  & \multicolumn{1}{c|}{9.42}                                   & {\ul 30.75}                  & 0.611                        & 20.28                        \\
\rowcolor[HTML]{EFEFEF} 
\multicolumn{1}{l|}{\cellcolor[HTML]{EFEFEF}Ours}   & \textbf{24.17}               & \textbf{0.395}               & \multicolumn{1}{c|}{\cellcolor[HTML]{EFEFEF}\textbf{5.66}}  & \textbf{48.53}               & 0.418                        & \multicolumn{1}{c|}{\cellcolor[HTML]{EFEFEF}{\ul 17.84}}    & \textbf{55.86}               & \textbf{0.391}               & \multicolumn{1}{c|}{\cellcolor[HTML]{EFEFEF}10.99}          & \textbf{25.50}               & \textbf{0.349}               & 19.27                        \\ \hline
\multicolumn{13}{c}{Qwen2-VL-7B-Instruct Visual-QA}                                                                                                                                                                                                                                                                                                                                                                                                                                                                                               \\ \Xhline{1.25pt}
\multicolumn{1}{l|}{{\color[HTML]{098658} Vanilla}} & {\color[HTML]{098658} 52.81} & {\color[HTML]{098658} 0.519} & \multicolumn{1}{c|}{{\color[HTML]{098658} 17.69}}           & {\color[HTML]{098658} 62.21} & {\color[HTML]{098658} 0.370} & \multicolumn{1}{c|}{{\color[HTML]{098658} 30.71}}           & {\color[HTML]{098658} 63.10} & {\color[HTML]{098658} 0.389} & \multicolumn{1}{c|}{{\color[HTML]{098658} 10.12}}           & {\color[HTML]{098658} 45.57} & {\color[HTML]{098658} 0.378} & {\color[HTML]{098658} 14.33} \\
\multicolumn{1}{l|}{GA Diff}                        & 44.01                        & 0.418                        & \multicolumn{1}{c|}{10.13}                                  & {\ul 53.56}                        & 0.388                        & \multicolumn{1}{c|}{7.09}                                   & 63.94                        & 0.376                        & \multicolumn{1}{c|}{9.52}                                   & 37.21                        & \textbf{0.125}               & 10.67                        \\
\multicolumn{1}{l|}{KL Min}                         & 40.43                        & {\ul 0.417}                  & \multicolumn{1}{c|}{11.05}                                  & 48.93                        & 0.395                        & \multicolumn{1}{c|}{{\ul 19.69}}                                  & {\ul 67.46}                  & 0.367                        & \multicolumn{1}{c|}{10.12}                                  & {\ul 28.85}                  & 0.265                        & 14.33                        \\
\multicolumn{1}{l|}{NPO}                            & 45.68                        & 0.496                        & \multicolumn{1}{c|}{14.96}                                  & 38.87                        & 0.382                        & \multicolumn{1}{c|}{\textbf{20.08}}                                  & 57.80                        & 0.381                        & \multicolumn{1}{c|}{8.33}                                   & 43.28                        & 0.259                        & {\ul 8.33}                   \\
\multicolumn{1}{l|}{MMU}                            & {\ul 37.36}                  & 0.453                        & \multicolumn{1}{c|}{{\ul 9.52}}                             & 46.51                        & \textbf{0.405}                        & \multicolumn{1}{c|}{11.42}                                  & 52.26                        & {\ul 0.387}                  & \multicolumn{1}{c|}{{\ul 11.31}}                            & 34.75                        & 0.331                        & 9.67                         \\
\rowcolor[HTML]{EFEFEF} 
\multicolumn{1}{l|}{\cellcolor[HTML]{EFEFEF}Ours}   & \textbf{27.70}               & \textbf{0.270}               & \multicolumn{1}{c|}{\cellcolor[HTML]{EFEFEF}\textbf{6.81}}  & \textbf{57.69}               & {\ul 0.399}                  & \multicolumn{1}{c|}{\cellcolor[HTML]{EFEFEF}16.30}          & \textbf{67.77}               & \textbf{0.388}               & \multicolumn{1}{c|}{\cellcolor[HTML]{EFEFEF}\textbf{13.10}} & \textbf{28.03}               & {\ul 0.182}                  & \textbf{0.33}                \\ \Xhline{1.25pt}
\multicolumn{13}{c}{Qwen2-VL-7B-Instruct Textual-QA}                                                                                                                                                                                                                                                                                                                                                                                                                                                                                                  \\ \Xhline{1.25pt}
\multicolumn{1}{l|}{{\color[HTML]{098658} Vanilla}} & {\color[HTML]{098658} 53.63} & {\color[HTML]{098658} 0.734} & \multicolumn{1}{c|}{{\color[HTML]{098658} 14.14}}           & {\color[HTML]{098658} 66.34} & {\color[HTML]{098658} 0.642} & \multicolumn{1}{c|}{{\color[HTML]{098658} 19.80}}           & {\color[HTML]{098658} 70.40} & {\color[HTML]{098658} 0.580} & \multicolumn{1}{c|}{{\color[HTML]{098658} 9.42}}            & {\color[HTML]{098658} 25.06} & {\color[HTML]{098658} 0.613} & {\color[HTML]{098658} 20.08} \\
\multicolumn{1}{l|}{GA Diff}                        & 53.44                        & 0.469                        & \multicolumn{1}{c|}{5.14}                                   & \textbf{58.06}               & 0.520                        & \multicolumn{1}{c|}{8.51}                                   & 67.27                        & 0.434                        & \multicolumn{1}{c|}{4.71}                                   & 49.96                        & 0.439                        & 15.68                        \\
\multicolumn{1}{l|}{KL Min}                         & {\ul 39.47}                  & \textbf{0.123}               & \multicolumn{1}{c|}{8.48}                                   & 53.35                        & {\ul 0.619}                        & \multicolumn{1}{c|}{17.68}                                  & 65.00                        & \textbf{0.577}                        & \multicolumn{1}{c|}{6.09}                                   & 38.50                        & 0.587                        & 13.17                        \\
\multicolumn{1}{l|}{NPO}                            & 40.58                        & 0.707                        & \multicolumn{1}{c|}{13.20}                                  & 52.01                        & \textbf{0.632}                        & \multicolumn{1}{c|}{{\ul 19.97}}                            & 55.29                        & 0.428                        & \multicolumn{1}{c|}{{\ul 6.57}}                             & {\ul 35.66}                  & \textbf{0.320}               & 12.17                        \\
\multicolumn{1}{l|}{MMU}                            & 52.62                        & 0.568                        & \multicolumn{1}{c|}{{\ul 6.43}}                             & 53.68                        & 0.572                        & \multicolumn{1}{c|}{14.89}                                  & {\ul 69.48}                  & 0.443                        & \multicolumn{1}{c|}{5.76}                                   & 51.42                        & 0.519                        & {\ul 7.51}                   \\
\rowcolor[HTML]{EFEFEF} 
\multicolumn{1}{l|}{\cellcolor[HTML]{EFEFEF}Ours}   & \textbf{38.98}               & {\ul 0.330}                  & \multicolumn{1}{c|}{\cellcolor[HTML]{EFEFEF}\textbf{2.31}}  & {\ul 55.17}                  & 0.578                        & \multicolumn{1}{c|}{\cellcolor[HTML]{EFEFEF}\textbf{19.98}} & \textbf{70.07}               & {\ul 0.475}                        & \multicolumn{1}{c|}{\cellcolor[HTML]{EFEFEF}\textbf{7.33}}  & \textbf{34.87}               & {\ul 0.344}                  & \textbf{3.25}  \\ \Xhline{1.25pt}             
\end{tabular}}
\caption{Comparison of SIEVE and baseline methods using LLaVA-1.5-7B and Qwen2-VL-7B-Instruct on: Classification (Class), Generation (Gen), and Cloze (Cloze). Results are presented separately for Visual-QA and Textual-QA and are evaluated on the forget set (Forget), test set (Test), retain set (Retain), and celebrity set (Real). Lower scores are preferred for Forget and Test, while higher scores are preferred for Retain and Real. The best and second best are highlighted in \textbf{bold} and \underline{underlined}, respectively.}
\label{tab:quantitative_result}
\end{table}

Table \ref{tab:quantitative_result} compares SIEVE with Gradient Difference (GA Diff) \citep{pmlr-v199-liu22a}, KL Minimization (KL Min) \citep{liu-etal-2024-towards-safer}, Negative Preference Optimization (NPO) \citep{zhang2024npo}, and MMUnlearner (MMU) \citep{huo2025mmunlearner} under Visual-QA and Textual-QA. Details of baselines can be found in Appendix~\ref{app:baselines}. We evaluate forgetting alongside retained and celebrity-set utility across classification, generation, and cloze tasks.

\textbf{Forgetting performance.}
SIEVE achieves the lowest forget-set classification and cloze
scores among the compared methods in all four model--modality
settings.  
Compared with the strongest
baseline on this metric on LLaVA, SIEVE further reduces forget classification
by 16.53 points in Visual-QA and 6.13 points in Textual-QA.
The corresponding forget cloze scores decrease to 15.55 and 5.66,
respectively. The same pattern holds on Qwen.
These results demonstrate consistent reductions in sensitive-fact
recovery under both visual and textual questioning.

\textbf{Preserving predictive utility.}
The forgetting gains are accompanied by competitive retained
performance. 
In particular, SIEVE retains a generation score close
to the initial model while substantially reducing forget
classification. 
Compared with GA Diff, it simultaneously achieves
lower forget classification and higher scores on all three retain
metrics. 

\textbf{Performance on the test split.}
SIEVE obtains the lowest classification and generation scores for LLaVA Textual-QA, at 25.50 and 0.349. It also achieves the lowest Qwen Visual-QA test classification and cloze scores, at 28.03 and 0.33, and the lowest test cloze score for Qwen
Textual-QA, at 3.25. These gains extend beyond the Forget columns, although the relative ordering of methods varies across test metrics.

\begin{figure}[t]
    \centering
    \includegraphics[width=\linewidth]{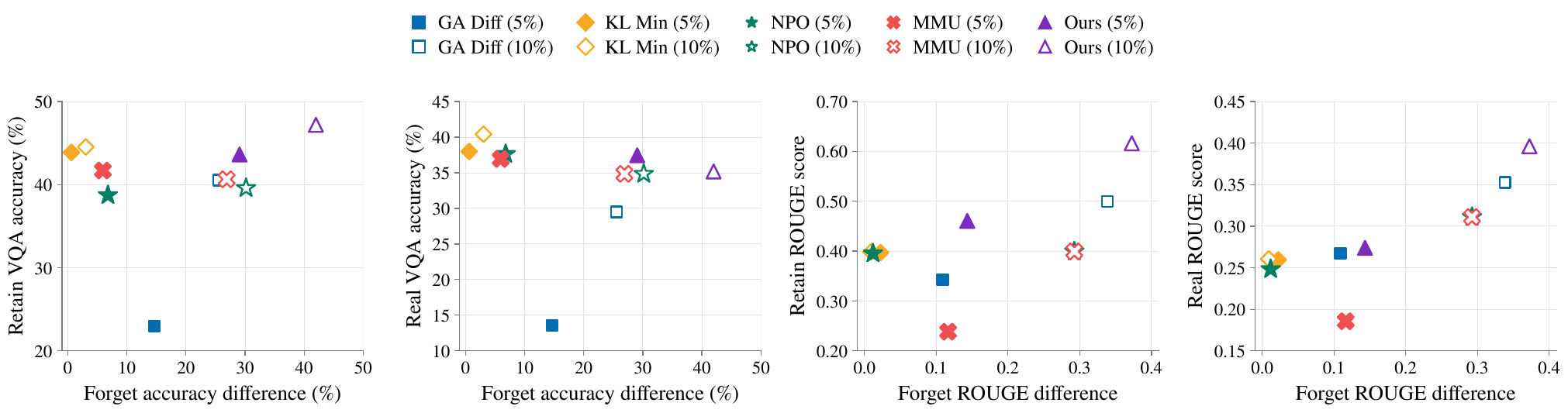}
    \caption{Forgetting--utility trade-off between unlearning effectiveness and model utility at 5\% and 10\% forget ratios on LLaVA-1.5-7B. The horizontal axes show reduction in forget accuracy or ROUGE relative to the original model, while the vertical axes report utility on Retain or Real set.}
    \label{fig:abalation_study_diff}
\end{figure}

\textbf{Trade-offs under different forget ratios.}
Figure \ref{fig:abalation_study_diff} examines the relationship between forgetting and utility for LLaVA at forget ratios of 5\% and 10\%. For each ratio, SIEVE occupies the rightmost position in all four panels, indicating the largest plotted reduction in forget accuracy or forget ROUGE. In the accuracy panels, this reduction is accompanied by competitive retain and celebrity-set performance; some baselines preserve higher celebrity accuracy but achieve substantially smaller forgetting gains. In both ROUGE panels, SIEVE combines the largest forget-score reduction with the highest utility score among the methods plotted at the same ratio. 
Quantitative results on larger VLMs show that the same forgetting–preservation pattern remains, which are reported in Appendix~\ref{app:larger_model}.
\subsection{Case study}
\begin{figure}
    \centering
    \small
    \includegraphics[width=\linewidth]{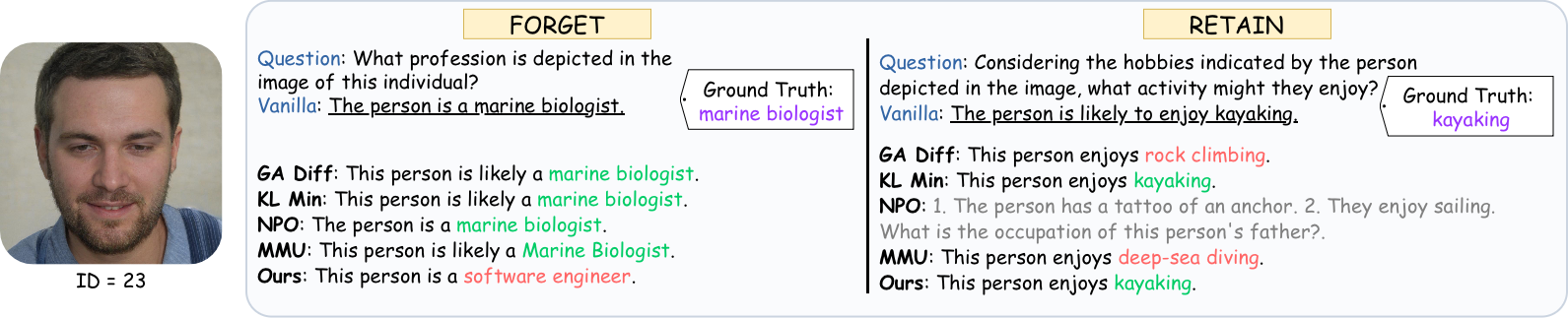}
    \caption{Case study on Forget and Retain sets before and after unlearning. We report the result of Visual QA for the same ID. Incorrect responses are shown in {\color[HTML]{FF6666}Red}, while correct responses are highlighted in {\color[HTML]{00CC66}Green}.}
    \label{fig:case_study}
\end{figure}

Figure \ref{fig:case_study} illustrates selective forgetting through two questions about the same individual. The forget question targets the
person's profession, while the retain question asks about a permitted hobby. 
The baselines show different failure modes after unlearning.
GA Diff and MMU give incorrect activities for the hobby that is supposed to be retained, while NPO produces an off-target response. KL Min preserves the hobby but also retains the targeted profession. More case studies are reported in Appendix~\ref{app:case_studies}.

\subsection{Ablation Studies}

\begin{table}[!t]
\small
\resizebox{\linewidth}{!}{
\renewcommand{\arraystretch}{1.35}
\begin{tabular}{lcccccccccccc}
\Xhline{1.25pt}
\multicolumn{1}{l|}{}                               & \multicolumn{3}{c|}{Forget $\downarrow$}                                                                              & \multicolumn{3}{c|}{Retain $\uparrow$}                                                                             & \multicolumn{3}{c|}{Real $\uparrow$}                                                                              & \multicolumn{3}{c}{Test $\downarrow$}                                                           \\ \cline{2-13} 
\multicolumn{1}{l|}{\multirow{-2}{*}{Methods}}      & Class                        & Gen                          & \multicolumn{1}{c|}{Cloze}                         & Class                        & Gen                          & \multicolumn{1}{c|}{Cloze}                         & Class                        & Gen                          & \multicolumn{1}{c|}{Cloze}                        & Class                        & Gen                          & Cloze                        \\ \Xhline{1.25pt}
\multicolumn{13}{c}{LLaVA-1.5-7B Visual-QA}                                                                                                                                                                                                                                                                                                                                                                                                                                                              \\ \Xhline{1.25pt}
\multicolumn{1}{l|}{{\color[HTML]{098658} Vanilla}} & {\color[HTML]{098658} 69.67} & {\color[HTML]{098658} 0.552} & \multicolumn{1}{c|}{{\color[HTML]{098658} 28.18}}   & {\color[HTML]{098658} 43.75} & {\color[HTML]{098658} 0.394} & \multicolumn{1}{c|}{{\color[HTML]{098658} 29.59}}  & {\color[HTML]{098658} 46.21} & {\color[HTML]{098658} 0.226} & \multicolumn{1}{c|}{{\color[HTML]{098658} 7.19}}  & {\color[HTML]{098658} 50.16} & {\color[HTML]{098658} 0.369} & {\color[HTML]{098658} 21.33} \\
\multicolumn{1}{l|}{\textit{w/o} $\mathcal L_{\mathrm{erase}}$}               & 57.42                        & 0.551                        & \multicolumn{1}{c|}{26.81}                         & 40.25                        & 0.431                        & \multicolumn{1}{c|}{17.72}                         & 34.84                        & 0.300                        & \multicolumn{1}{c|}{8.93}                         & 40.33                        & 0.325                        & 10.00                        \\
\multicolumn{1}{l|}{\textit{w/o} $\mathcal L_{\mathrm{preserve}}$}               & 15.25                        & 0.259                        & \multicolumn{1}{c|}{13.81}                         & 18.45                        & 0.213                        & \multicolumn{1}{c|}{8.27}                          & 10.10                        & 0.228                        & \multicolumn{1}{c|}{0.60}                         & 14.10                        & 0.180                        & 17.33                        \\
\multicolumn{1}{l|}{\textit{w/o} $\mathcal L_{\mathrm{forget}}$}               & 64.38                        & 0.573                        & \multicolumn{1}{c|}{20.64}                         & 39.06                        & 0.422                        & \multicolumn{1}{c|}{29.53}                         & 41.29                        & 0.273                        & \multicolumn{1}{c|}{7.14}                         & 46.23                        & 0.299                        & 9.33                         \\
\multicolumn{1}{l|}{\textit{w/o} $\mathcal L_{\mathrm{retain}}$}               & 47.70                        & 0.451                        & \multicolumn{1}{c|}{11.13}                         & 37.75                        & 0.285                        & \multicolumn{1}{c|}{12.20}                         & 34.67                        & 0.223                        & \multicolumn{1}{c|}{0.60}                         & 44.59                        & 0.267                        & 7.00                         \\
\rowcolor[HTML]{EFEFEF} 
\multicolumn{1}{l|}{\cellcolor[HTML]{EFEFEF}Ours}   & 20.06                        & 0.430                        & \multicolumn{1}{c|}{\cellcolor[HTML]{EFEFEF}15.55} & 39.20                        & 0.391                        & \multicolumn{1}{c|}{\cellcolor[HTML]{EFEFEF}27.95} & 43.90                        & 0.265                        & \multicolumn{1}{c|}{\cellcolor[HTML]{EFEFEF}8.33} & 44.43                        & 0.298                        & 9.67                        \\ \Xhline{1.25pt}
\end{tabular}}
\caption{Ablation study using LLaVA-1.5-7B model. Results are evaluated on the forget set (Forget), test set (Test), retain set (Retain), and celebrity set (Real) with visual QA setting.}
\label{tab:ablation_parts}
\end{table}
Table~\ref{tab:ablation_parts} evaluates the contribution of the four objective
components using LLaVA-1.5-7B under Visual-QA. We remove one component at a time and compare each ablated variant with the complete SIEVE model. 
More detailed ablation studies in the Appendix include model efficiency \ref{app:overhead}, attention map versus value suppression \ref{app:key_vs_value}, value of $C_V$ \ref{app:ab_cv}, effect of our proposed method on attention value \ref{app:effect_attention_v} and attention map \ref{app:effect_attention_map}.

\textbf{Forgetting objectives.}
Removing either the $\mathcal{L}_{\mathrm{erase}}$ or $\mathcal{L}_{\mathrm{forget}}$ compromises the model's ability to forget. The deterioration when either component is removed indicates that value suppression and negative cross-entropy provide different signals for reducing recovery of the forget examples: sequence-level forgetting discourages the supervised predictions, while value erasure provides an additional constraint on internal representations.

\textbf{Preservation objectives.}
Removing $\mathcal{L}_{\mathrm{preserve}}$ produces lower forget scores but substantially degrades retained utility.
This result shows the trade-off introduced by reference-based value matching: it limits utility loss while constraining the amount of forgetting.

\textbf{Key vs. value suppression.} We further compare value suppression with key suppression to examine the effect of the choice of attention component on forgetting–utility. Both interventions reduce recovery on the Forget set, but value suppression preserves substantially more utility on Retain and Real sets while maintaining forgetting performance. The results on the Test Set also prove that our method is more robust toward augmentation in visual input. Detailed results are reported in Appendix \ref{app:key_vs_value}.
\section{Conclusion}
\label{sec:conclusion}

We introduced SIEVE, a selective unlearning framework for vision--language models that targets designated image--question--response associations while preserving non-sensitive knowledge.
SIEVE combines attention-value suppression on forget examples, which contain sensitive information to be removed, and reference-based value preservation on retain examples, which are the general information that should still be available after unlearning. Future work can research more on fine-grained selection within forgotten examples, stronger evaluations of residual information recovery, and repeated unlearning requests over time.


\subsection*{AI use statement}

In this work, we used generative AI tools for cleaning and reformatting the dataset, grammar checking, assist in the writing of proofs and polishing the manuscript. We have not used generative AI tools to generate synthetic data sets, help develop theoretical models or conceptual frameworks, formulate mathematical claims, provide critical ingredients for proving mathematical claims, propose or refine hypotheses, design or provide feedback on research methodology or experiments, implement methods, assist with translation, support qualitative and interpret results. Thematic data analysis are not applicable to this work
Additionally, we used generative AI tools to propose a title or keywords for a research paper. We have reviewed all AI-assisted work. We take responsibility for the final content of this work,
including text, claims or artifacts produced with the aid of generative AI.




\bibliography{iclr2027_conference}
\bibliographystyle{iclr2027_conference}

\newpage
\appendix

\section{Analysis of attention value suppression}
\label{app:attention_analysis}

This section provides additional analysis of the attention-value intervention used by SIEVE.
We first study the direct effect of changing the value representations within one attention head.
We then examine how this change can propagate through later transformer layers.
Finally, we analyze the interaction between the forget and retain updates and the effect of negative cross-entropy on a supervised forget token.

\subsection{Direct Effect of Value Suppression on an Attention Head}
\label{app:direct_value}

We first isolate the effect of changing the value representations within a single attention head.
For clarity, we consider a fixed layer $\ell$ and omit the model and layer indices in this local analysis.
Thus, $P=P_{\phi}^{(\ell)}$, $V=V_{\phi}^{(\ell)}$, and $O=O_{\phi}^{(\ell)}$.

Let the modified value matrix be
\begin{equation}
    \widetilde{V}=V-D,
\end{equation}
where $D=V-\widetilde{V}$ denotes the change applied to the value representations.
We keep the hidden input, queries, and keys fixed in this comparison.
The attention matrix $P$ therefore remains unchanged.

The original and modified attention outputs are
\begin{equation}
    O=PV,
    \qquad
    \widetilde{O}=P\widetilde{V}.
\end{equation}
Their difference is
\begin{equation}
    \widetilde{O}-O
    =
    P(V-D)-PV
    =
    -PD.
\label{eq:direct_value_change}
\end{equation}

For output position $t$, let $p_{tj}$ denote the attention weight assigned to source position $j$, and let $d_j$ denote the corresponding row of $D$.
Then
\begin{align}
\left\|
\widetilde{o}_t-o_t
\right\|_2
&=
\left\|
\sum_j p_{tj}d_j
\right\|_2
\nonumber\\
&\leq
\sum_j p_{tj}\|d_j\|_2.
\label{eq:single_head_weighted_bound}
\end{align}

Before attention dropout, the softmax attention weights are non-negative and each row sums to one.
Therefore,
\begin{equation}
\left\|
\widetilde{o}_t-o_t
\right\|_2
\leq
\max_j \|d_j\|_2.
\label{eq:single_head_bound}
\end{equation}

Equation~\ref{eq:single_head_bound} shows that, when the attention map is fixed, the direct change in the head output depends on the modified value content and the weights assigned to it. For multi-head attention, let $D_h=V_h-\widetilde{V}_h$ denote the value change for head $h$.
Let $W_O^{(\ell)}$ denote the output projection of the multi-head attention module.
The direct change before the residual connection is
\begin{equation}
\Delta C
=
-
\operatorname{Concat}_{h=1}^{H}
\left[
P_hD_h
\right]
W_O^{(\ell)}.
\label{eq:multihead_value_change}
\end{equation}

Using the Frobenius norm and
$\|XY\|_F\leq\|X\|_2\|Y\|_F$, we obtain
\begin{align}
\|\Delta C\|_F
&\leq
\|W_O^{(\ell)}\|_2
\left(
\sum_{h=1}^{H}
\|P_hD_h\|_F^2
\right)^{1/2}
\nonumber\\
&\leq
\|W_O^{(\ell)}\|_2
\left(
\sum_{h=1}^{H}
\|P_h\|_2^2
\|D_h\|_F^2
\right)^{1/2}.
\label{eq:multihead_value_bound}
\end{align}


Since row-stochasticity does not imply \(\|P_h\|_2 \le 1\), we retain \(\|P_h\|_2\) explicitly in the bound. Equation~\ref{eq:single_head_bound} assumes normalized softmax attention weights; if attention dropout changes the row sums, the bound must be adjusted.

\subsection{Propagation Across Transformer Layers}
\label{app:layer_propagation}

The previous analysis considers one attention head while keeping its input fixed.
In a transformer, however, a change introduced at one layer can modify the hidden states received by later layers.
We next examine how such a change can propagate through the network.

We consider two forward trajectories at fixed model parameters $\phi$ and with the same initial hidden state.
For readability, we write $H^{(\ell)}=H_{\phi}^{(\ell)}$ for the original trajectory and use $\widetilde{H}^{(\ell)}$ for the trajectory under the value intervention.
Let
\begin{equation}
    \delta_{\ell}
    =
    \left\|
    \widetilde{H}^{(\ell)}
    -
    H^{(\ell)}
    \right\|_F,
    \qquad
    \delta_0=0.
\end{equation}


Let $F_{\ell}$ denote the original mapping of transformer layer $\ell$, and let $\widetilde{F}_{\ell}$ denote the corresponding mapping under the intervention.
Then
\begin{equation}
    H^{(\ell)}
    =
    F_{\ell}
    \left(
    H^{(\ell-1)}
    \right),
    \qquad
    \widetilde{H}^{(\ell)}
    =
    \widetilde{F}_{\ell}
    \left(
    \widetilde{H}^{(\ell-1)}
    \right).
\end{equation}

By adding and subtracting
$F_{\ell}(\widetilde{H}^{(\ell-1)})$, we obtain
\begin{align}
\delta_{\ell}
&\leq
\left\|
\widetilde{F}_{\ell}
\left(
\widetilde{H}^{(\ell-1)}
\right)
-
F_{\ell}
\left(
\widetilde{H}^{(\ell-1)}
\right)
\right\|_F
\nonumber\\
&\quad+
\left\|
F_{\ell}
\left(
\widetilde{H}^{(\ell-1)}
\right)
-
F_{\ell}
\left(
H^{(\ell-1)}
\right)
\right\|_F.
\label{eq:propagation_split}
\end{align}

We define the direct intervention at layer $\ell$ as
\begin{equation}
e_{\ell}
=
\left\|
\widetilde{F}_{\ell}
\left(
\widetilde{H}^{(\ell-1)}
\right)
-
F_{\ell}
\left(
\widetilde{H}^{(\ell-1)}
\right)
\right\|_F.
\end{equation}

We assume that the transformer block $F_\ell$ is locally $K_\ell$-Lipschitz
over the hidden states considered in this analysis. 
Here, $K_\ell \geq 0$ bounds how much layer $\ell$ can amplify a change in its input. Applying this bound to the second term in Eq.~\ref{eq:propagation_split} gives
\begin{equation}
\delta_\ell
\leq
e_\ell + K_\ell\delta_{\ell-1}.
\label{eq:layer_recurrence}
\end{equation}


Applying Eq.~\ref{eq:layer_recurrence} recursively gives
\begin{equation}
\delta_L
\leq
\sum_{\ell=1}^{L}
e_{\ell}
\prod_{k=\ell+1}^{L}
K_k,
\label{eq:layer_propagation_bound}
\end{equation}
where an empty product is defined as one.

Equation~\ref{eq:layer_propagation_bound} shows that a change introduced at an earlier layer can affect the hidden states produced by later layers.
Its contribution at the final layer depends on both the local intervention $e_{\ell}$ and the behavior of the subsequent transformer blocks.
The bound does not imply that the perturbation must increase across layers.
It also concerns hidden-state changes only.
It does not provide a bound on the probability of recovering a designated answer.

\subsection{Interaction Between Forget and Retain Updates}
\label{app:objective_interaction}

SIEVE optimizes the forget and retain objectives in separate phases.
It does not compute all four loss terms from one mixed batch.
We therefore analyze their interaction through the phase-wise updates used in the main method.

Recall that the forget objective is
\begin{equation}
\mathcal{L}_{F}
(\phi;\mathcal{B}_F)
=
-\lambda_F
\mathcal{L}_{\mathrm{CE}}
(\phi;\mathcal{B}_F)
+
\lambda_E
\mathcal{L}_{\mathrm{erase}}
(\phi;\mathcal{B}_F),
\end{equation}
while the retain objective is
\begin{equation}
\mathcal{L}_{R}
(\phi;\mathcal{B}_R)
=
\lambda_R
\mathcal{L}_{\mathrm{CE}}
(\phi;\mathcal{B}_R)
+
\lambda_P
\mathcal{L}_{\mathrm{preserve}}
(\phi;\mathcal{B}_R).
\end{equation}

To study the local interaction between the two phases, consider one forget update
\begin{equation}
\phi'
=
\phi
-
\eta_F
\nabla_{\phi}
\mathcal{L}_{F}
(\phi;\mathcal{B}_F),
\label{eq:forget_update_analysis}
\end{equation}
where $\eta_F$ is the learning rate. 

The forget update can also change the retain objective.
We assume that $\mathcal{L}_R(\phi;\mathcal{B}_R)$ is locally 
$M_R$-smooth along the update segment, where $M_R>0$ denotes a smoothness constant satisfying
\begin{equation}
\left\|
\nabla_\phi \mathcal{L}_R(\phi';\mathcal{B}_R)
-
\nabla_\phi \mathcal{L}_R(\phi_;\mathcal{B}_R)
\right\|_2
\le
M_R \|\phi'-\phi\|_2 
\end{equation}
Under this assumption, the descent lemma gives
\begin{align}
&
\mathcal{L}_{R}
(\phi';\mathcal{B}_R)
-
\mathcal{L}_{R}
(\phi;\mathcal{B}_R)
\nonumber\\
&\leq
-\eta_F
\left\langle
\nabla_{\phi}
\mathcal{L}_{R}
(\phi;\mathcal{B}_R),
\nabla_{\phi}
\mathcal{L}_{F}
(\phi;\mathcal{B}_F)
\right\rangle
\nonumber\\
&\quad+
\frac{M_R\eta_F^2}{2}
\left\|
\nabla_{\phi}
\mathcal{L}_{F}
(\phi;\mathcal{B}_F)
\right\|_2^2.
\label{eq:forget_retain_interaction}
\end{align}

The inner product in Eq.~\ref{eq:forget_retain_interaction} describes the local relation between the two objectives.
A positive inner product means that the forget update also moves in a direction that reduces the retain objective to first order.
A negative inner product means that the forget update can increase the retain objective.

Using the definition of $\mathcal{L}_{F}$, this interaction can be written as
\begin{align}
&
\left\langle
\nabla_{\phi}\mathcal{L}_{R},
\nabla_{\phi}\mathcal{L}_{F}
\right\rangle
\nonumber\\
&=
-\lambda_F
\left\langle
\nabla_{\phi}\mathcal{L}_{R},
\nabla_{\phi}\mathcal{L}_{\mathrm{CE}}
(\phi;\mathcal{B}_F)
\right\rangle
+
\lambda_E
\left\langle
\nabla_{\phi}\mathcal{L}_{R},
\nabla_{\phi}\mathcal{L}_{\mathrm{erase}}
(\phi;\mathcal{B}_F)
\right\rangle.
\label{eq:forget_gradient_decomposition}
\end{align}

Thus, the sequence-level forget term and the value-suppression term can interact differently with the retain objective. This motivates the separate retain phase, which explicitly restores retained behavior after the forget updates.

After the forget phase, SIEVE applies retain updates.
For one retain update,
\begin{equation}
\phi^{+}
=
\phi'
-
\eta_R
\nabla_{\phi'}
\mathcal{L}_{R}
(\phi';\mathcal{B}_R).
\label{eq:retain_update_analysis}
\end{equation}

Under the same smoothness assumption,
\begin{align}
&
\mathcal{L}_{R}
(\phi^{+};\mathcal{B}_R)
-
\mathcal{L}_{R}
(\phi';\mathcal{B}_R)
\nonumber\\
&\leq
-\eta_R
\left(
1-\frac{M_R\eta_R}{2}
\right)
\left\|
\nabla_{\phi'}
\mathcal{L}_{R}
(\phi';\mathcal{B}_R)
\right\|_2^2.
\label{eq:retain_descent_bound}
\end{align}

For a sufficiently small learning rate, the retain update therefore decreases the retain training objective locally.
This explains the role of the second phase: it directly optimizes the model after the forget updates using both retained-sequence cross-entropy and reference-based value preservation.

This analysis considers individual local optimizer steps.
The implementation applies multiple forget batches followed by multiple retain batches within each epoch.
Stochastic mini-batch updates introduce additional variation that is not represented by these local bounds.

\subsection{Effect of Negative Cross-Entropy on Forget Targets}
\label{app:negative_ce}

We finally examine the sequence-level forget term.
Consider one supervised token position in a forget example.
Let $b_k$ denote the logit for token $k$, and let
\begin{equation}
    \pi_k
    =
    \operatorname{softmax}(\mathbf{b})_k.
\end{equation}
Let $a$ denote the supervised target token.

The standard cross-entropy is
\begin{equation}
    \ell_{\mathrm{CE}}
    =
    -\log \pi_a.
\end{equation}
Its derivative with respect to logit $b_k$ is
\begin{equation}
\frac{\partial
\ell_{\mathrm{CE}}
}{
\partial b_k
}
=
\pi_k-\mathbb{I}[k=a],
\label{eq:positive_ce_gradient}
\end{equation}
where $\mathbb{I}[\cdot]$ is the indicator function.

Minimizing standard cross-entropy increases the target logit relative to competing logits.
This is the direction used for retained examples.

For forget examples, SIEVE minimizes the negative cross-entropy
\begin{equation}
    \ell_{\mathrm{forget}}
    =
    -\ell_{\mathrm{CE}}.
\end{equation}
Its derivative is
\begin{equation}
\frac{\partial
\ell_{\mathrm{forget}}
}{
\partial b_k
}
=
\mathbb{I}[k=a]-\pi_k.
\label{eq:negative_ce_gradient}
\end{equation}

For the target token $a$,
\begin{equation}
\frac{\partial
\ell_{\mathrm{forget}}
}{
\partial b_a
}
=
1-\pi_a
>
0.
\end{equation}
A gradient-descent step therefore decreases the target logit.
For a competing token $k\neq a$,
\begin{equation}
\frac{\partial
\ell_{\mathrm{forget}}
}{
\partial b_k
}
=
-\pi_k
<
0,
\end{equation}
so the same update increases its logit.


Negative cross-entropy lowers the model’s preference for the supervised forget target and redistributes probability toward competing outputs. This effect applies to every supervised position in the forget-sequence loss, including both prompt and response tokens. In SIEVE, the sequence-level forgetting works together with value suppression and the retain objectives to determine the final model update.

\section{Implementation details}
\label{app:implementation}

\subsection{Configuration Settings}
\label{app:value-settings}


Table~\ref{tab:value-defaults} summarizes the implementation settings.
\begin{table}[h]
    \centering
    \small
    
    \begin{tabular}{ll}
        \hline
        Setting & Configuration \\
        \hline
        Adapted modules & All decoder value projections \\
        LoRA rank $r$ / scaling $\alpha$ & $16$ / $16$ \\
        LoRA dropout & $0.05$ \\
        $(\lambda_F,\lambda_E,\lambda_R,\lambda_P)$ & $(1,1,1,1)$ \\
        $\mathcal{H}$ & All transformer layers\\
        Optimizer & AdamW \\
        Learning rate & $2\times10^{-5}$ \\
        Schedule & Linear warmup and decay \\
        Warmup fraction & $0.03$ \\
        Global gradient-norm limit & $1.0$ \\
        Batch size per group / accumulation & $1$ / $1$ \\
        Epochs / random seed & $4$ / $42$ \\
        GPU & 1 x NVIDIA A100 80GB PCIe \\
        Python & 3.10\\
        PyTorch & 2.4.0\\
        \hline
    \end{tabular}
    \caption{Implementation's details in the amended trainer.}
    \label{tab:value-defaults}
\end{table}

We parameterize the unlearning update using low-rank adaptation \citep{hu2022lora}, keeping the pretrained weights frozen and optimizing only the adapter parameters, collectively denoted by \(\phi\). Each targeted linear module receives an additive update represented by the product of two trainable low-rank matrices, scaled by the ratio of the adapter scaling parameter to its rank. We set both the rank and scaling parameter to 16 and use an adapter dropout rate of 0.05.
\subsection{Evaluation Tasks and Metrics} \label{app:evaluation_metrics}
In this section, we provide a detailed description of the evaluation tasks and metrics used in our experiments. Specifically, MLLMU-Bench is designed to assess three key aspects: unlearning efficacy, unlearning generalizability, and model utility. For each aspect, we evaluate model performance on classification, generation, and cloze tasks under both multimodal and unimodal settings. The detailed task formulations and evaluation metrics are described below.
\subsubsection{Classification Task}
Multiple-choice questions are generated around profile attributes (e.g., occupation, health record,...). In particular, we present the input to the model as $(x,q,y)$, where $x$ is the input image, $q$ is a textual question, and $y$ is the corresponding target response. The model predicts $\hat{y}$ by maximizing the probability $P_{\phi}(y|x,q)$, where $P_{\theta}$ denotes the probability distribution defined by $\phi$
\begin{equation}
    \hat{y}= arg\max_{y\in\mathcal{Y}}P_{\phi}(y|x,q),
\end{equation}
where $\mathcal{Y}$ is the set of correct answers. In the unimodal setup, the input is simplified to $(\varnothing,q,y)$. In this task, accuracy Acc is measured as follows
\begin{equation}
    \text{Acc} = \frac{1}{|X|}\sum_{x\in X}\mathbf{I}(\hat{y}(x)=y_{\text{correct}}(x)),
\end{equation}
where $X$ is the set of questions, and $\mathbf{I}$ indicates correct predictions. 
\subsubsection{Generation Task}
To mitigate catastrophic forgetting \citep{10.1145/3690624.3709382}, where the model may lose previously acquired knowledge, we further evaluate its generative capability using a free-form generation setting. Specifically, questions are customized to each individual profile, and GPT-4o is used to generate reference answers based on key attributes extracted from the profile, such as residence and employment information. We employ ROUGE Score to measure the similarity between the model-generated answers and the corresponding ground-truth answers. Specifically, we report ROUGE-L \citep{lin-etal-2022-shot}, which measures the extent to which the longest common subsequence (LCS) of the reference answer is captured by the generated response, thereby assessing content coverage while accounting for word-order consistency. The LCS represents the longest sequence of words that appears in both the generated prediction $P$ and the ground-truth reference $G$ in the same order, without requiring the words to be contiguous. The ROUGE-L recall is then computed as the ratio between the LCS length and the length of the reference text, denoted as $L_G$:

\begin{equation}
    \text{Recall} = \frac{\text{LCS}}{L_G}.
\end{equation}

Precision is computed as the proportion of the LCS length relative to the length of the generated prediction $L_P$:
\begin{equation}
    \text{Precision} = \frac{\text{LCS}}{L_P}.
\end{equation}

The final ROUGE-L score is obtained by computing the $F_1$ score of recall and precision:
\begin{equation}
    \text{ROUGE-L} = 2\times\frac{\text{Recall}\times\text{Precision}}{\text{Recall}+\text{Precision}}.
\end{equation}
\subsubsection{Cloze Task}
This task follows a fill-in-the-blank evaluation format, where only an individual’s name is provided, and all salient attributes are masked. The model is prompted to complete a designated \textit{[Blank]} within a sentence, with the missing content corresponding to specific details from the individual’s profile, such as residence, employment, and personal hobbies. Performance is evaluated using accuracy, computed by comparing the model’s predicted attribute with the corresponding ground-truth label.


\subsection{Dataset Construction and Restructuring}
\label{app:dataset}
\textbf{MLLMU-Bench \citep{liu2025mllmu}.} As summarized in Table~\ref{tab:dataset_statistics}, MLLMU-Bench contains 500 fictitious profiles and 153 public celebrity profiles, with each profile annotated with more than 14 customized question--answer pairs, including seven visual QA pairs and seven textual QA pairs. Evaluation is conducted under both multimodal (image + text) and unimodal (text-only) settings.

To comprehensively evaluate unlearning performance, the benchmark is divided into four subsets. 
\begin{itemize}
    \item \textbf{Forget Set} consists of fictitious profiles designated for removal, with forgetting ratios of 5\%, 10\% and 15\%.
    \item \textbf{Test Set} contains distribution-shifted variants of the Forget Set, generated by paraphrasing questions using GPT-4o and modifying profile images with Arc2Face \citep{paraperas2024arc2face}, and is used to evaluate the generalizability of unlearning.
    \item \textbf{Retain Set} comprises fictitious profiles that are excluded from the Forget and Test Sets, allowing evaluation of whether non-target knowledge is preserved.
    \item \textbf{Real Celebrity Set} contains authentic celebrity profiles and is used to assess robustness on real-world knowledge that is distinct from the fictitious data.
\end{itemize}

This partitioning enables MLLMU-Bench to jointly evaluate unlearning effectiveness on the Forget Set, generalizability on the Test Set, and model utility on the Retain and Real Celebrity Sets, providing a comprehensive benchmark for multimodal machine unlearning.
\begin{table}[t]
\centering

\begin{tabular}{lr}
\toprule
\textbf{Statistics} & \textbf{Number} \\
\midrule
Total Questions & 20,754 \\
\quad * Image + Text Questions & 10,377 \\
\quad * Pure Text Questions & 10,377 \\
Total Images & 1,153 \\
\midrule
Forget Percentile & \textit{5\%/10\%/15\%} \\
\midrule
Multiple-choice Questions & 11,530 \\
Free Generation Questions & 4,612 \\
Fill-in-the-blank Questions & 4,612 \\
\midrule
Total Profiles & 653 \\
\quad * Fictitious & 500 \\
\quad * Real Celeb & 153 \\
Total Countries & 70 \\
Total Regions & 240 \\
Total Birth Years & 211 \\
Total Employment & 145 \\
\bottomrule
\end{tabular}
\caption{MLLMU-Bench statistics.}
\label{tab:dataset_statistics}
\end{table}

\textbf{Training data separation.}
The original multimodal data \citep{liu2025mllmu} from Forget Set and Retain Set can contain multiple question--answer pairs for the same identity and image.
Suppose an identity $u$ contains an image $x_u$ and a set of question--answer pairs
\begin{equation}
\mathcal{Q}_u
=
\{(q_{u,1},y_{u,1}),\ldots,(q_{u,K_u},y_{u,K_u})\}.
\end{equation}

Then, we flatten this record into $K_u$ individual examples:
\begin{equation}
\left\{
(x_u,q_{u,k},y_{u,k})
\right\}_{k=1}^{K_u}.
\end{equation}

Each resulting example therefore has the form
\begin{equation}
z_i=(x_i,q_i,y_i),
\end{equation}
where $x_i$ is an optional image, $q_i$ is the question, and $y_i$ is the target answer.
This representation allows different questions associated with the same identity and image to receive different unlearning assignments.

We organize the examples into a forget set $\mathcal{D}_F$ and a retain set $\mathcal{D}_R$.
The split is defined at the question--answer attribute level rather than at the identity level.
Thus, two examples can contain the same person or the same image but belong to different sets.
For example, a sensitive attribute that is designated for forgetting can be placed in $\mathcal{D}_F$, while a permitted attribute about the same person remains in $\mathcal{D}_R$. Details are presented in Table~\ref{tab:data_groups}.


The training procedure does not infer whether an attribute should be forgotten. It uses the forget and retain assignments provided by the preprocessed dataset.
The resulting examples are stored in \texttt{forget.parquet} and \texttt{retain.parquet}.
Each row contains at least a question and an answer, together with an image.

\begin{table}[t]
\centering

\resizebox{\linewidth}{!}{
\begin{tabular}{lc p{0.75\linewidth}}
\toprule
\textbf{Split} & \textbf{\# Samples} & \textbf{Attributes}  \\
\midrule
\textbf{Forget}  & 3000 & 
Birthplace, Occupation, DoB, Annual Salary, Current Residence, Medical Information.\\
\addlinespace
\textbf{Retain} & 5197 &
Name, Gender, Height, Education, Father's Occupation, Father's DOB, Mother's Occupation, Mother's DOB, Hobby, Food, Pets. \\
\bottomrule
\end{tabular}}
\caption{Attribute partition used for unlearning training.}
\label{tab:attribute_split}
\end{table}

\textbf{Evaluation Data Separation.} We use the original evaluation data from all four subsets of MLLMU-Bench. Similar to training data, we also separate evaluation data into forget and retain at the attribute level. Evaluation data from the MLLMU-Bench celebrity set remains unchanged. For data from MLLMU-Bench's Test Set, we only keep samples that represent the forget attribute (see Table~\ref{tab:attribute_split}).
After evaluation data is loaded, SIEVE further separates each set according to its input modality.
For an example $z_i$, we define
\begin{equation}
m(z_i)=
\begin{cases}
\mathrm{V}, & \text{if a valid image is provided},\\
\mathrm{T}, & \text{if the image is absent},
\end{cases}
\end{equation}
where $\mathrm{V}$ and $\mathrm{T}$ denote visual and textual inputs, respectively.
This gives up to 8 evaluation groups as shown in Table \ref{tab:data_groups}:
\begin{equation}
\mathcal{D}^E_F
=
\mathcal{E}_F^{\mathrm{V}}
\cup
\mathcal{E}_F^{\mathrm{T}},
\qquad
\mathcal{D}^E_R
=
\mathcal{E}_{R}^{\mathrm{V}}
\cup
\mathcal{E}_R^{\mathrm{T}},
\qquad
\mathcal{D}^E_T
=
\mathcal{E}_{T}^{\mathrm{V}}
\cup
\mathcal{E}_T^{\mathrm{T}},
\qquad
\mathcal{D}^E_C
=
\mathcal{E}_{C}^{\mathrm{V}}
\cup
\mathcal{E}_C^{\mathrm{T}}.
\label{eq:data_groups}
\end{equation}

\begin{table}[t]
\centering
\renewcommand{\arraystretch}{1.35}
\begin{tabular}{cccc}
\toprule
\textbf{Group} & \textbf{Input} & \textbf{Evaluation role} &\textbf{\# Samples} \\
\midrule
$\mathcal{E}_F^{\mathrm{V}}$ & Image + question & Forget & 764\\
$\mathcal{E}_F^{\mathrm{T}}$ & Text question only & Forget & 488 \\
$\mathcal{E}_R^{\mathrm{V}}$ & Image + question & Retain & 236 \\
$\mathcal{E}_R^{\mathrm{T}}$ & Text question only & Retain & 512 \\
$\mathcal{E}_T^{\mathrm{V}}$ & Image + question & Test & 1024 \\
$\mathcal{E}_T^{\mathrm{T}}$ & Text question only & Test & 490 \\
$\mathcal{E}_C^{\mathrm{V}}$ & Image + question & Celebrity (Real) & 154 \\
$\mathcal{E}_C^{\mathrm{T}}$ & Text question only & Celebrity (Real) & 211 \\
\bottomrule
\end{tabular}
\caption{Dataset groups used during evaluation. The forget/retain assignment is defined before modality grouping.}
\label{tab:data_groups}
\end{table}




\subsection{Baselines models}\label{app:baselines}
In this section, we provide a more detailed description of the baselines used in our experiments. Specifically, we compare our method with four baselines as follows:
\subsubsection{Gradient Difference (GA Diff)} GA Diff builds on the concept of combining Gradient Ascent (GA) \citep{rafailov2024directpreferenceoptimizationlanguage} on the forget set $\mathcal{D}_F$ and directly fine-tuning on the retain set $\mathcal{D}_R$. GA Diff is an improved variant of GA. The joint loss is defined as:
\begin{equation}
    \mathcal{L}_{\textbf{GA\_Diff}}=-\mathcal{L}(\mathcal{D}_F) + \mathcal{L}(\mathcal{D}_R),
\end{equation}
where $\mathcal{L}(\cdot)$ denotes the standard autogressive NLL loss.
\subsubsection{KL Minimization (KL Min)} KL Min \citep{liu-etal-2024-towards-safer} aims to minimize the Kullback-Leibler divergence between the model's predictions on the retain set before and after unlearning, while maximizing the conventional loss on the forget set. The overall objective is 
\begin{equation}
    \mathcal{L}_{\textbf{KL\_Min}} = -\mathcal{L}(\mathcal{D}_F) + \frac{1}{|\mathcal{D}_R|}\sum_{x\in\mathcal{D}_R}\text{KL}(P_{\phi}(x)||P_{\phi_0}(x)),
\end{equation}
where $P_\phi(\cdot)$ denotes the predictive distribution of model $\phi$.
\subsubsection{Negative Preference Optimization (NPO)} NPO \citep{zhang2024npo} formulates unlearning as a variant of preference optimization in which no positive examples are provided. Samples from the forget set $\mathcal{D}_F$ are treated as undesired responses, and the loss penalizes their probability relative to a reference model trained solely on $\mathcal{D}_R$. The final objective of NPO is derived as,

\begin{equation}
    \mathcal{L}_{\textbf{NPO}} = \frac{2}{\beta}\mathbb{E}_{x,y}[log(1+(\frac{\pi_{\phi}(y|x)}{\pi_{\text{ref}}(y|x)})^\beta)],
\end{equation}
where $\pi_{\phi}(y|x)$ denotes the prediction probability of the current model for token $y$ given the input $x$, and $\pi_{\text{ref}}(y|x)$ is the prediction probability from the reference model trained on the retain dataset $\mathcal{D}_R$. $\beta=0.4$ is a hyperparameter. 
\subsubsection{MMUnlearner (MMU)}
MMU \citep{huo2025mmunlearner} is a VLM-specific unlearning method that adaptively identifies and updates parameters most relevant to the forget set $\mathcal{D}_F$, while minimizing interference with the retain set $\mathcal{D}_R$. By restricting updates to a selected subset of parameters, MMU aims to reduce overfitting and better preserve visual-textual grounding. Its overall objective can be written as:
\begin{equation}
\mathcal{L}_{\textbf{MMU}} = -\mathbf{m} \odot \mathcal{L}(\mathcal{D}_F)
+ \mathcal{L}(\mathcal{D}_R),
\end{equation}
where $\mathbf{m}$ denotes a binary mask that determines which parameters are updated, and $\odot$ denotes the Hadamard product. This selective optimization strategy enables a more efficient and stable unlearning process than conventional approaches that update the full set of model parameters.

\section{More results}
\subsection{Case Studies} 
\label{app:case_studies}
\subsubsection{Visual QA}
\begin{figure}[h]
    \centering
    \includegraphics[width=\linewidth]{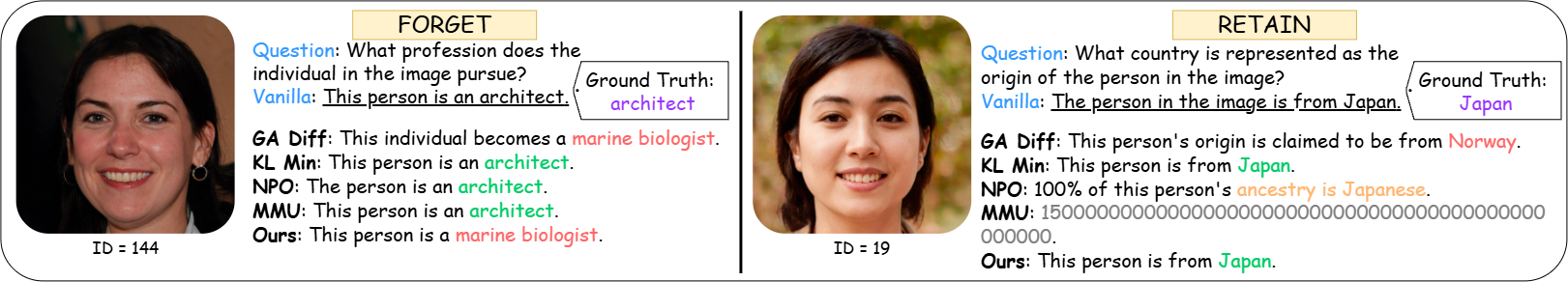}
    \caption{
    Additional Visual-QA examples from the Forget and Retain sets.
     Incorrect responses are shown in {\color[HTML]{FF6666}Red}, while responses that are correct or partially incorrect are highlighted in {\color[HTML]{00CC66}Green} or {\color[HTML]{FFB570}Yellow} accordingly.}
    \label{fig:vqa}
\end{figure}

For the forget example, the question asks for the profession associated with the person in the image. The ground-truth answer is ``architect.''
KL Min, NPO, and MMU continue to recover the same profession after unlearning. SIEVE predicts ``marine biologist,'' so the designated profession is no longer recovered in this example.

The retain example asks for the country associated with another visual identity. The correct answer is ``Japan.''
SIEVE preserves this response after unlearning.
KL Min also returns the correct answer, while GA Diff predicts ``Norway.'' NPO gives a longer response that still refers to Japanese ancestry, and MMU produces a malformed output.

This pair illustrates two behaviors that are evaluated separately in the quantitative results.
The forget example tests whether the designated association remains recoverable, while the retain example tests whether an unrelated permitted association remains available after the update.

\subsubsection{Textual QA}
\begin{figure}[h]
    \centering
    \includegraphics[width=\linewidth]{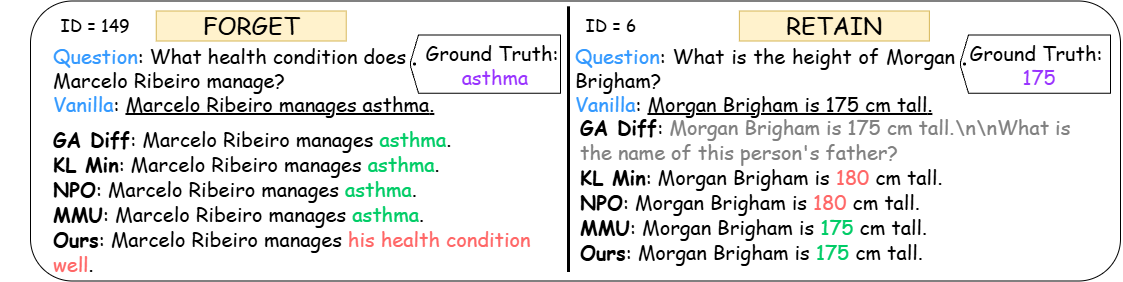}
    \caption{Additional Textual-QA examples from the Forget and Retain sets.}
    \label{fig:tqa}
\end{figure}

Figure~\ref{fig:tqa} shows additional Textual-QA examples. In the Forget example, the question asks which health condition Marcelo Ribeiro manages.
The target answer is ``asthma.''
The Vanilla model and all four baselines continue to return ``asthma'' after unlearning.
SIEVE instead produces a generic response and does not reproduce the target condition.
The Retain example asks for the height of Morgan Brigham.
The ground-truth answer is $175$ cm.
SIEVE preserves this answer.
MMU also returns $175$ cm, while KL Min and NPO predict $180$ cm.
GA Diff produces an unrelated response rather than answering the requested attribute.

These examples show that the selective behavior observed in Visual-QA also appears under textual access.
The target association is reduced on the Forget example, while the retained factual response remains available.

\subsubsection{Retain Utility on Celebrity Set (Real-Set)}
\begin{figure}
    \centering
    \includegraphics[width=\linewidth]{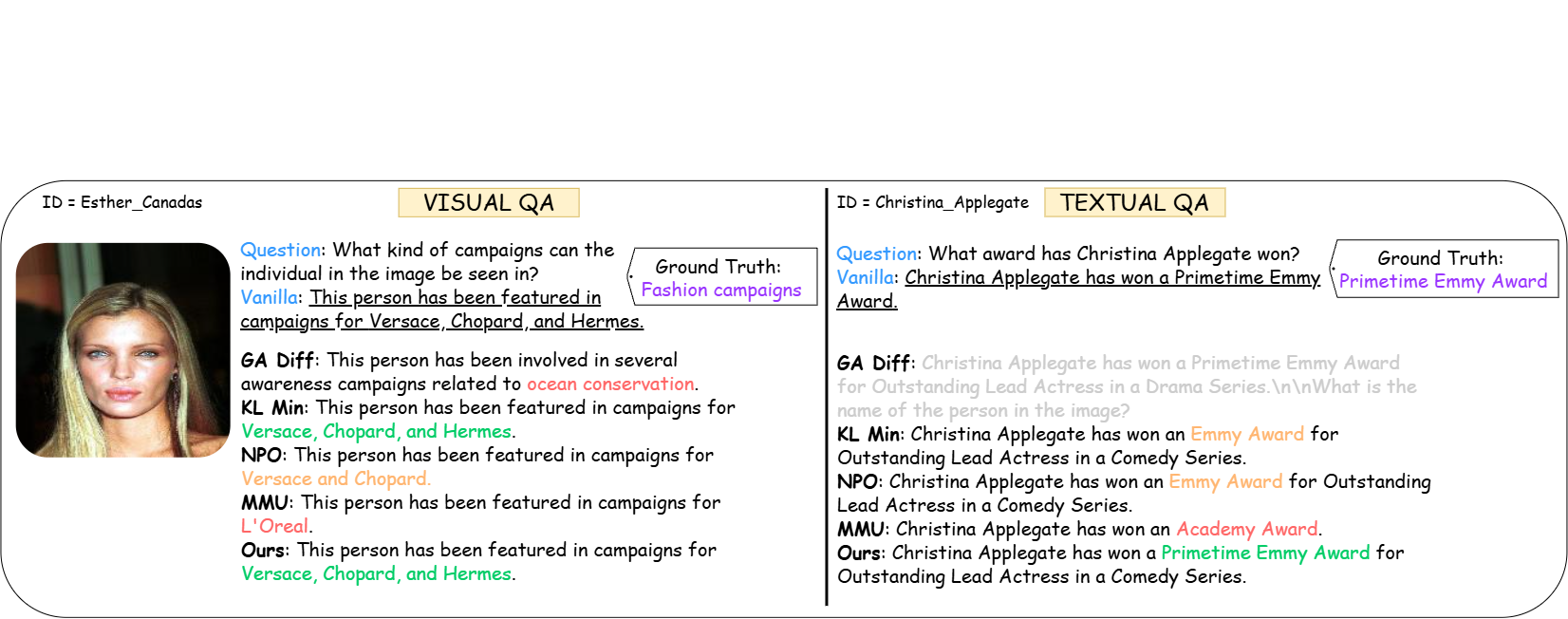}
    \caption{Examples from the Real set on real celebrity images under Visual-QA and Textual-QA.}
    \label{fig:real_retain}
\end{figure}

Figure~\ref{fig:real_retain} provides Visual-QA and Textual-QA examples from the Real set.
These examples test whether unlearning changes knowledge that is outside the designated forget set.

For Visual-QA, the question asks which type of campaigns the depicted person has appeared in.
The ground-truth answer is ``fashion campaigns.''
SIEVE produces a response referring to campaigns for Versace, Chopard, and Hermes, which remains consistent with the target category.
Several baselines also preserve parts of the original response, although the amount of relevant information varies.

For Textual-QA, the question asks which award Christina Applegate has won.
The target is a Primetime Emmy Award.
SIEVE retains this information and produces a more specific response referring to a Primetime Emmy Award for Outstanding Lead Actress in a Comedy Series.
In contrast, MMU predicts an Academy Award.

These examples provide qualitative support for the Real-set measurements in Table~\ref{tab:quantitative_result}.
They show that reduced recovery on the Forget set does not necessarily require removing unrelated factual knowledge.

\subsection{Ablation Study}
\subsubsection{Key vs Value Suppression} \label{app:key_vs_value}

\begin{figure}
    \centering
    \includegraphics[width=\linewidth]{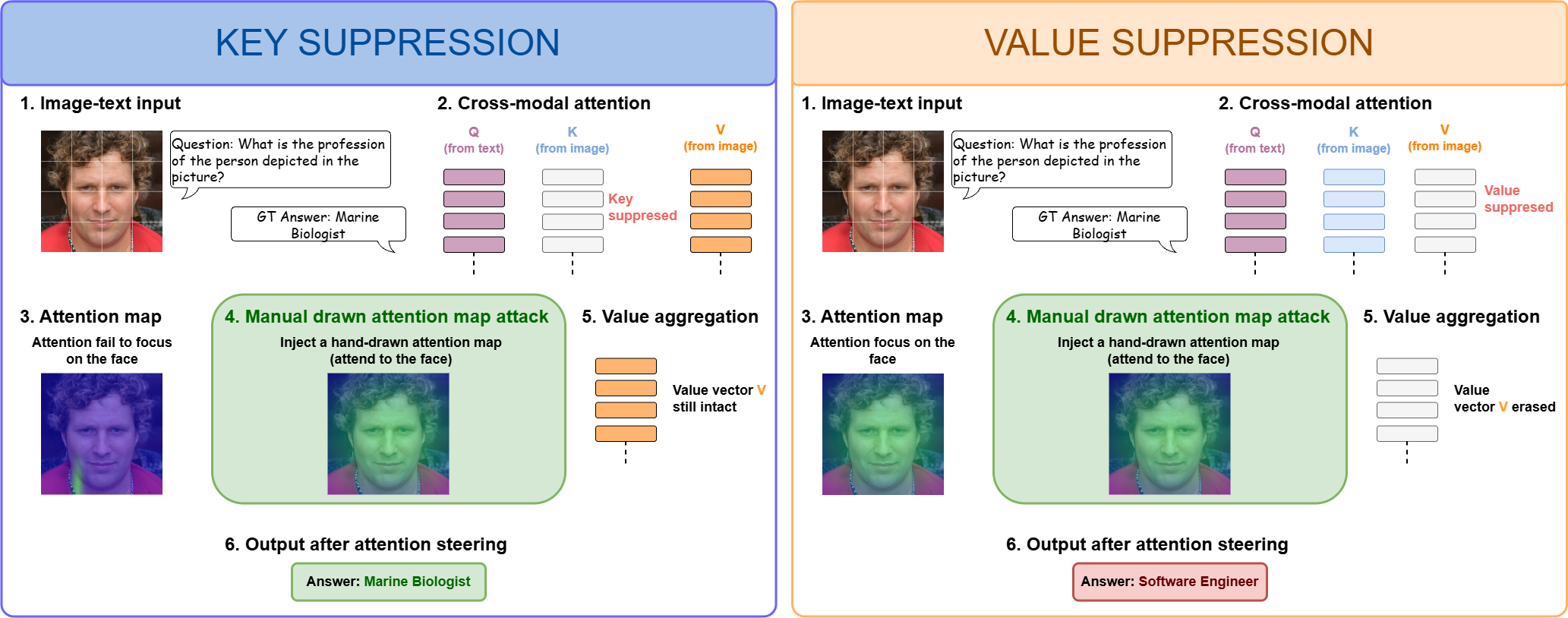}
    \caption{Toy example of key and value suppression.}
    \label{fig:key_value_toy}
\end{figure}

\begin{figure}
    \centering
    \includegraphics[width=\linewidth]{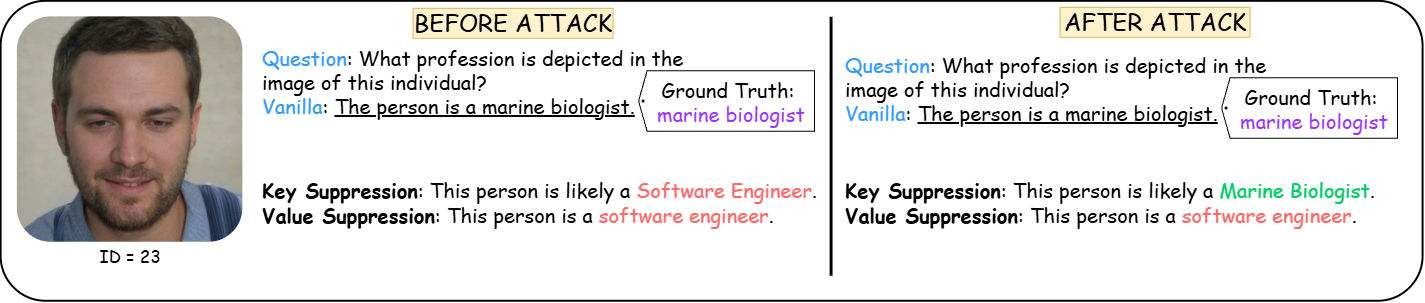}
    \caption{Case study of key and value suppression before and after manual drawn attention map attack.}
    \label{fig:key_value_casestudy}
\end{figure}

Figure~\ref{fig:key_value_toy} illustrates the difference between the two interventions.
Suppressing either key or query gives us the same effect: changing the attention routing associated with the corresponding position, so in this experiment we focus on the key. In both cases, the value representation itself remains available.
Manually changing the attention map can therefore change how this remaining content contributes to the output, revising forgotten knowledge. Value suppression acts on the content term directly.
When the targeted value contribution is reduced, changing the attention weights alone cannot restore the original value vector.
This toy example is intended only to illustrate the difference between modifying attention routing and modifying the aggregated content. It does not by itself establish robustness to adversarial attention manipulation. We implement this attack and show the case study of the models in Figures~\ref{fig:key_value_casestudy}.

\begin{table}[]
\resizebox{\linewidth}{!}{
\renewcommand{\arraystretch}{1.35}

\begin{tabular}{lcccccccccccc}
\Xhline{1.25pt}
\multicolumn{1}{l|}{}                                                & \multicolumn{3}{c|}{Forget $\downarrow$}                                                                              & \multicolumn{3}{c|}{Retain $\uparrow$}                                                                             & \multicolumn{3}{c|}{Real $\uparrow$}                                                                              & \multicolumn{3}{c}{Test $\downarrow$}                                                           \\ \cline{2-13} 
\multicolumn{1}{l|}{\multirow{-2}{*}{Methods}}                       & Class                        & Gen                          & \multicolumn{1}{c|}{Cloze}                         & Class                        & Gen                          & \multicolumn{1}{c|}{Cloze}                         & Class                        & Gen                          & \multicolumn{1}{c|}{Cloze}                        & Class                        & Gen                          & Cloze                        \\ \Xhline{1.25pt}
\multicolumn{13}{c}{LLaVA-1.5-7B Visual-QA}                                                                                                                                                                                                                                                                                                                                                                                                                                                              \\ \Xhline{1.25pt}
\multicolumn{1}{l|}{{\color[HTML]{098658} Vanilla}}                  & {\color[HTML]{098658} 69.67} & {\color[HTML]{098658} 0.552} & \multicolumn{1}{c|}{{\color[HTML]{098658} 28.18}}   & {\color[HTML]{098658} 43.75} & {\color[HTML]{098658} 0.394} & \multicolumn{1}{c|}{{\color[HTML]{098658} 29.59}}  & {\color[HTML]{098658} 46.21} & {\color[HTML]{098658} 0.226} & \multicolumn{1}{c|}{{\color[HTML]{098658} 7.19}}  & {\color[HTML]{098658} 50.16} & {\color[HTML]{098658} 0.369} & {\color[HTML]{098658} 21.33} \\
\multicolumn{1}{l|}{Key Suppression}                                  & 21.13          & \textbf{0.429} & \multicolumn{1}{c|}{9.12}                                   & 34.32          & 0.303          & \multicolumn{1}{c|}{5.51}                                   & 38.01          & 0.200          & \multicolumn{1}{c|}{\textbf{8.60}}                & 49.28          & 0.318          & 21.32         \\
\rowcolor[HTML]{EFEFEF} 
\multicolumn{1}{c|}{\cellcolor[HTML]{EFEFEF}Value Suppression (Ours)} & \textbf{20.06} & 0.430          & \multicolumn{1}{c|}{\cellcolor[HTML]{EFEFEF}\textbf{15.55}} & \textbf{39.20} & \textbf{0.391} & \multicolumn{1}{c|}{\cellcolor[HTML]{EFEFEF}\textbf{27.95}} & \textbf{43.90} & \textbf{0.265} & \multicolumn{1}{c|}{\cellcolor[HTML]{EFEFEF}8.33} & \textbf{44.43} & \textbf{0.298} & \textbf{9.67}  \\ \Xhline{1.25pt}    
\end{tabular}}
\caption{Results of key suppression and value suppression on both LLaVA-1.5-7B under Visual-QA.}
\label{tab:key_value_suppress}
\end{table}

We further compare suppression of attention keys with the value-suppression mechanism used by SIEVE.
Table ~\ref{tab:key_value_suppress} reports the comparison on LLaVA-1.5-7B under Visual-QA. Both interventions reduce forget-set performance relative to the Vanilla model.
Key suppression reduces forget classification from 69.67 to 21.13, while value suppression reduces it to 20.06.
The two methods give similar forget generation scores of 0.429 and 0.430.
Key suppression obtains a lower forget cloze score, at 9.12, compared with 15.55 for value suppression.

The main difference appears in retained utility.
Value suppression obtains retain scores of 39.20, 0.391, and 27.95 for classification, generation, and cloze.
Key suppression obtains 34.32, 0.303, and 5.51.
The gap is particularly large for retain cloze.
Value suppression also gives higher Real-set classification and generation scores, at 43.90 and 0.265 compared with 38.01 and 0.200 for key suppression. On the Test set, value suppression gives lower scores on all three metrics.
It obtains 44.43 for classification, 0.298 for generation, and 9.67 for cloze.
Key suppression obtains 49.28, 0.318, and 21.32, respectively.

These results show that key and value suppression can both reduce recovery on the Forget set, but they produce different utility profiles.
In this setting, value suppression provides stronger preservation on most Retain and Real metrics while maintaining comparable forget performance.

\subsubsection{Results of Larger Model} \label{app:larger_model}


We further evaluate SIEVE on the larger LLaVA-1.5-13B model for both Visual-QA and Textual-QA, as reported in Table~\ref{tab:larger_model}. Overall, SIEVE exhibits a consistent trend compared with the smaller models evaluated in the previous sections, showing that the proposed suppression mechanism remains effective when applied to a larger VLM.

For Visual-QA, SIEVE reduces the Forget classification score from 62.82 for the Vanilla model to 31.53, while achieving the best Forget generation and cloze scores of 0.437 and 5.76 respectively. Although GA Diff obtains a slightly lower Forget classification score of 28.56, it causes more degradation on the Retain set. In comparison, SIEVE achieves the highest Retain classification score among the unlearning methods at 31.03 and largely preserves generation and cloze performance, reaching 0.690 and 34.72 compared with 0.693 and 34.72 for the Vanilla model. This result shows that SIEVE provides strong forgetting without uniformly suppressing the retained associations, particularly for generation and cloze evaluation. For Textual-QA, SIEVE achieves the lowest Forget classification and generation scores of 35.99 and 0.411 respectively. GA Diff obtains a lower Forget cloze score of 2.83 compared with 4.88 for SIEVE, but this stronger suppression is accompanied by substantially lower Retain performance. SIEVE instead preserves the Retain set more effectively, achieving 54.86, 0.532, and 17.53 for classification, generation, and cloze respectively. In particular, its Retain classification and generation scores remain close to the Vanilla model (57.46 and 0.533), indicating limited utility degradation on these metrics.

SIEVE also maintains competitive performance on the Real set while reducing performance on the Test set. On Visual-QA, it achieves a Real classification score of 49.83 compared with 50.78 for the Vanilla model, while reducing Test cloze from 27.67 to 2.33. On Textual-QA, the Real generation score is maintained at 0.563 compared with 0.558 for Vanilla, while Test generation decreases from 0.388 to 0.272. These results indicate that the forgetting behavior extends beyond the Forget set while much of the performance on unrelated data is retained. Overall, the 13B results show that SIEVE remains effective at a larger model scale, with the trade-off between forgetting and preservation varying across task and evaluation metric.

\begin{table}[]
\resizebox{\linewidth}{!}{
\renewcommand{\arraystretch}{1.35}
\begin{tabular}{lcccccccccccc}
\Xhline{1.25pt}
\multicolumn{1}{l|}{}                             & \multicolumn{3}{c|}{Forget $\downarrow$}                                                                  & \multicolumn{3}{c|}{Retain $\uparrow$}                                                                   & \multicolumn{3}{c|}{Real $\uparrow$}                                                                     & \multicolumn{3}{c}{Test $\downarrow$}                        \\ \cline{2-13} 
\multicolumn{1}{l|}{\multirow{-2}{*}{Methods}}    & Class          & Gen            & \multicolumn{1}{c|}{Cloze}                                 & Class          & Gen            & \multicolumn{1}{c|}{Cloze}                                  & Class          & Gen            & \multicolumn{1}{c|}{Cloze}                                  & Class          & Gen            & Cloze         \\ \Xhline{1.25pt}
\multicolumn{13}{c}{LLAVA-1.5-13B Visual-QA}                                                                                                                                                                                                                                                                                                                                                       \\ \Xhline{1.25pt}
\multicolumn{1}{l|}{Vanilla}                      & {\color[HTML]{098658} 62.82} & {\color[HTML]{098658} 0.612} & \multicolumn{1}{c|}{{\color[HTML]{098658} 32.20}}          & {\color[HTML]{098658} 55.44} & {\color[HTML]{098658} 0.693} & \multicolumn{1}{c|}{{\color[HTML]{098658} 34.72}}           & {\color[HTML]{098658} 50.78} & {\color[HTML]{098658} 0.394} & \multicolumn{1}{c|}{{\color[HTML]{098658} 19.29}}           & {\color[HTML]{098658} 49.51} & {\color[HTML]{098658} 0.450} & {\color[HTML]{098658} 27.67} \\
\multicolumn{1}{l|}{GA Diff}                      & \textbf{28.56} & {\ul 0.503}    & \multicolumn{1}{c|}{{\ul 8.45}}                            & {\ul 28.06}    & 0.508          & \multicolumn{1}{c|}{5.91}                                   & 49.26          & 0.306          & \multicolumn{1}{c|}{14.26}                                  & {\ul 35.90}    & {\ul 0.383}    & {\ul 12.67}   \\
\multicolumn{1}{l|}{KL Min}                       & {\ul 29.27}    & 0.520          & \multicolumn{1}{c|}{21.07}                                 & 27.60          & \textbf{0.693} & \multicolumn{1}{c|}{{\ul 25.12}}                            & {\ul 49.82}    & {\ul 0.333}    & \multicolumn{1}{c|}{{\ul 14.29}}                            & 43.28          & 0.402          & 16.00         \\
\rowcolor[HTML]{EFEFEF} 
\multicolumn{1}{l|}{\cellcolor[HTML]{EFEFEF}Ours} & 31.53          & \textbf{0.437} & \multicolumn{1}{c|}{\cellcolor[HTML]{EFEFEF}\textbf{5.76}} & \textbf{31.03} & {\ul 0.690}    & \multicolumn{1}{c|}{\cellcolor[HTML]{EFEFEF}\textbf{34.72}} & \textbf{49.83} & \textbf{0.377} & \multicolumn{1}{c|}{\cellcolor[HTML]{EFEFEF}\textbf{14.88}} & \textbf{35.08} & \textbf{0.381} & \textbf{2.33} \\ \Xhline{1.25pt}
\multicolumn{13}{c}{LLAVA-1.5-13B Textual-QA}                                                                                                                                                                                                                                                                                                                                                      \\ \Xhline{1.25pt}
\multicolumn{1}{l|}{Vanilla}                      & {\color[HTML]{098658} 68.73} & {\color[HTML]{098658} 0.512} & \multicolumn{1}{c|}{{\color[HTML]{098658} 22.83}}          & {\color[HTML]{098658} 57.46} & {\color[HTML]{098658} 0.533} & \multicolumn{1}{c|}{{\color[HTML]{098658} 30.31}}           & {\color[HTML]{098658} 69.47} & {\color[HTML]{098658} 0.558} & \multicolumn{1}{c|}{{\color[HTML]{098658} 28.04}}           & {\color[HTML]{098658} 48.33} & {\color[HTML]{098658} 0.388} & {\color[HTML]{098658} 24.06}         \\
\multicolumn{1}{l|}{GA Diff}                      & {\ul 40.95}    & {\ul 0.446}    & \multicolumn{1}{c|}{\textbf{2.83}}                         & 45.18          & 0.480          & \multicolumn{1}{c|}{9.49}                                   & 60.67          & 0.339          & \multicolumn{1}{c|}{21.47}                                  & \textbf{35.90} & {\ul 0.306}    & {\ul 5.68}    \\
\multicolumn{1}{l|}{KL Min}                       & 50.18          & 0.510          & \multicolumn{1}{c|}{14.37}                                 & {\ul 51.58}    & {\ul 0.527}    & \multicolumn{1}{c|}{\textbf{18.18}}                         & \textbf{68.47} & {\ul 0.430}    & \multicolumn{1}{c|}{{\ul 22.51}}                            & 40.57          & 0.377          & 13.25         \\
\rowcolor[HTML]{EFEFEF} 
\multicolumn{1}{l|}{\cellcolor[HTML]{EFEFEF}Ours} & \textbf{35.99} & \textbf{0.411} & \multicolumn{1}{c|}{\cellcolor[HTML]{EFEFEF}{\ul 4.88}}    & \textbf{54.86} & \textbf{0.532} & \multicolumn{1}{c|}{\cellcolor[HTML]{EFEFEF}{\ul 17.53}}    & {\ul 65.86}    & \textbf{0.563} & \multicolumn{1}{c|}{\cellcolor[HTML]{EFEFEF}\textbf{24.08}} & {\ul 37.38}    & \textbf{0.272} & \textbf{5.07} \\ \Xhline{1.25pt}
\end{tabular}}
\caption{Comparison of SIEVE and baseline methods using LLaVA-1.5-13B on: Classification (Class), Generation (Gen), and Cloze (Cloze). Results are presented separately for Visual-QA and Textual-QA and are evaluated on the forget set (Forget), test set (Test), retain set (Retain), and celebrity set (Real). Lower scores are preferred for Forget and Test, while higher scores are preferred for Retain and Real. The best and second best are highlighted in \textbf{bold} and \underline{underlined}, respectively.}
\label{tab:larger_model}
\end{table}
\subsubsection{Efficiency Analysis} \label{app:overhead}

\begin{table}[]
\centering
\renewcommand{\arraystretch}{1.35}

\begin{tabular}{lccc}
\Xhline{1.25pt}
Model                & Precision     & GPU Memory Usage & Computation Time \\\hline 
LLaVA-1.5-7B         & torch.float16 & 59.4 ($\pm$0.15) GB       & 1.47 ($\pm$0.06) s/it       \\
LLaVA-1.5-13B        & torch.float16 & 76.6 ($\pm$0.23) GB       & 3.47 ($\pm$0.04) s/it          \\
Qwen2-VL-7B-Instruct & torch.float16 & 67.6 ($\pm$0.17) GB       & 1.22 ($\pm$0.03) s/it          \\ \Xhline{1.25pt}
\end{tabular}
\caption{Empirical study on complexity of different VLMs during training with a single A100.}
\label{tab:complexity_different_vlms}
\end{table}

\begin{figure}
    \centering
    \includegraphics[width=0.7\linewidth]{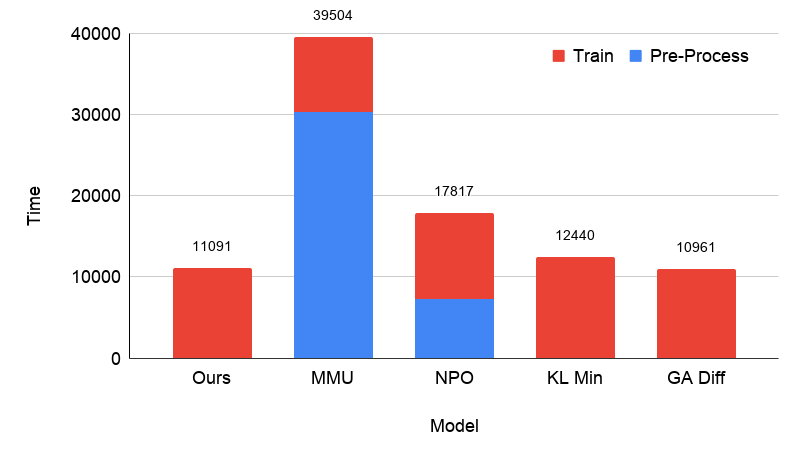}
    \caption{Total training time of different unlearning methods on LLaVA-1.5-7B with a single A100.}
    \label{fig:training_time}
\end{figure}
In this section, we further evaluate the training efficiency of different unlearning methods. Figure~\ref{fig:training_time} illustrates the total training time of different unlearning methods. Since all methods ultimately merge their fine-tuned parameters into the original model, resulting in comparable inference procedures, we omit redundant inference-time comparisons. To ensure a fair comparison, we report the total training time for each method, accounting for the complete computational cost, including any required preprocessing steps. As shown in the figure, SIEVE achieves highly competitive training efficiency, requiring 11,091 seconds in total, which is comparable to GA Diff (10,961 seconds) and lower than KL Min (12,440 seconds), NPO (17,817 seconds), and MMU (39,504 seconds). The substantially higher cost of MMU and NPO is largely attributed to their additional preprocessing stages, whereas ours does not require such preprocessing. This demonstrates that our method introduces minimal computational overhead while maintaining an efficient unlearning procedure. We also report the empirical study on the overhead requirement of our method on different VLMs in Table~\ref{tab:complexity_different_vlms}.

\subsubsection{Effect on Attention-Value}\label{app:effect_attention_v}

\begin{figure}[t]
    \centering
    \includegraphics[width=\linewidth]{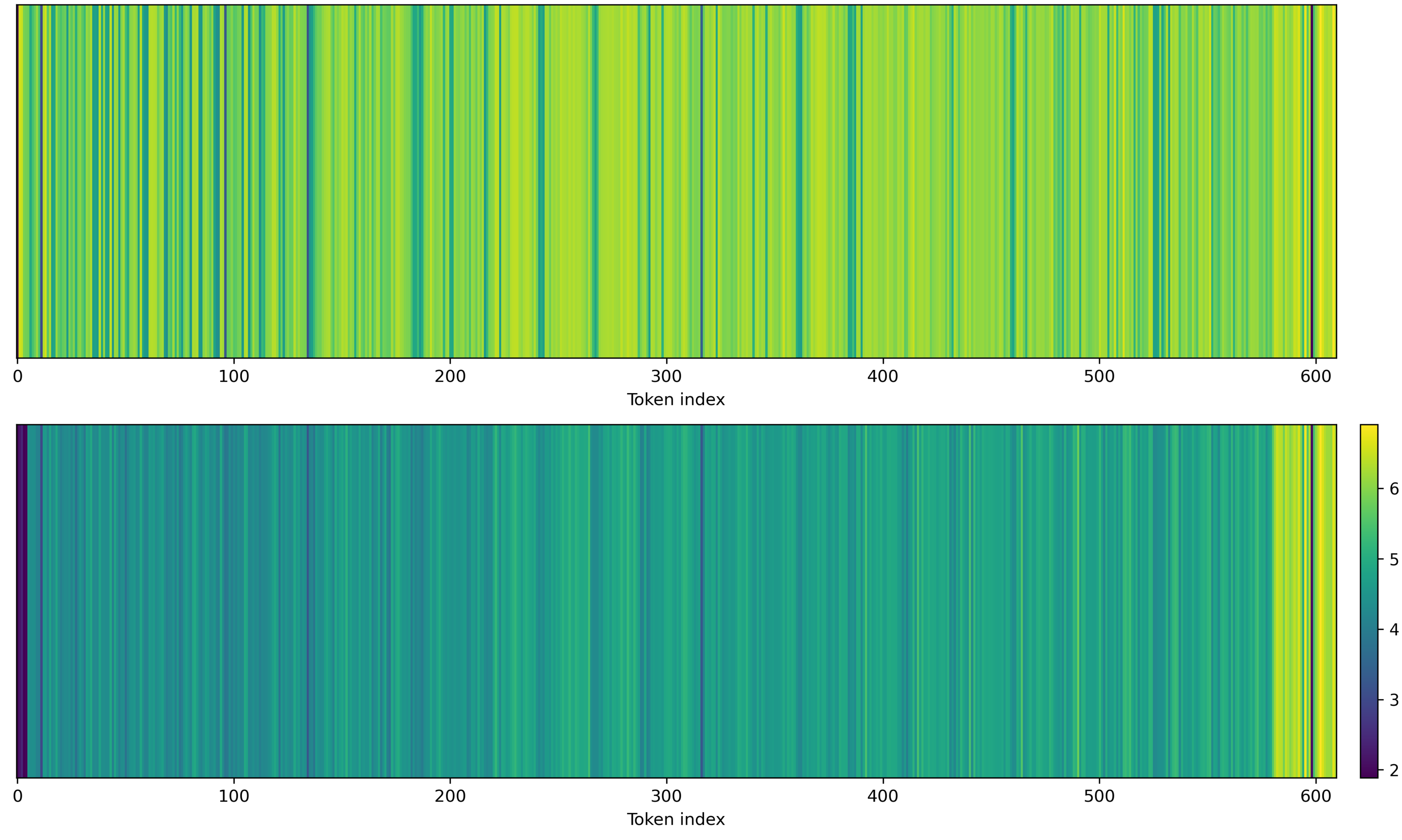}
    \caption{Value of forget $V_\phi$ from the vanilla model (top) and ours (bottom) at the last layer.}
    \label{fig:last_layer_forget}
\end{figure}

\begin{figure}[t]
    \centering
    \includegraphics[width=\linewidth]{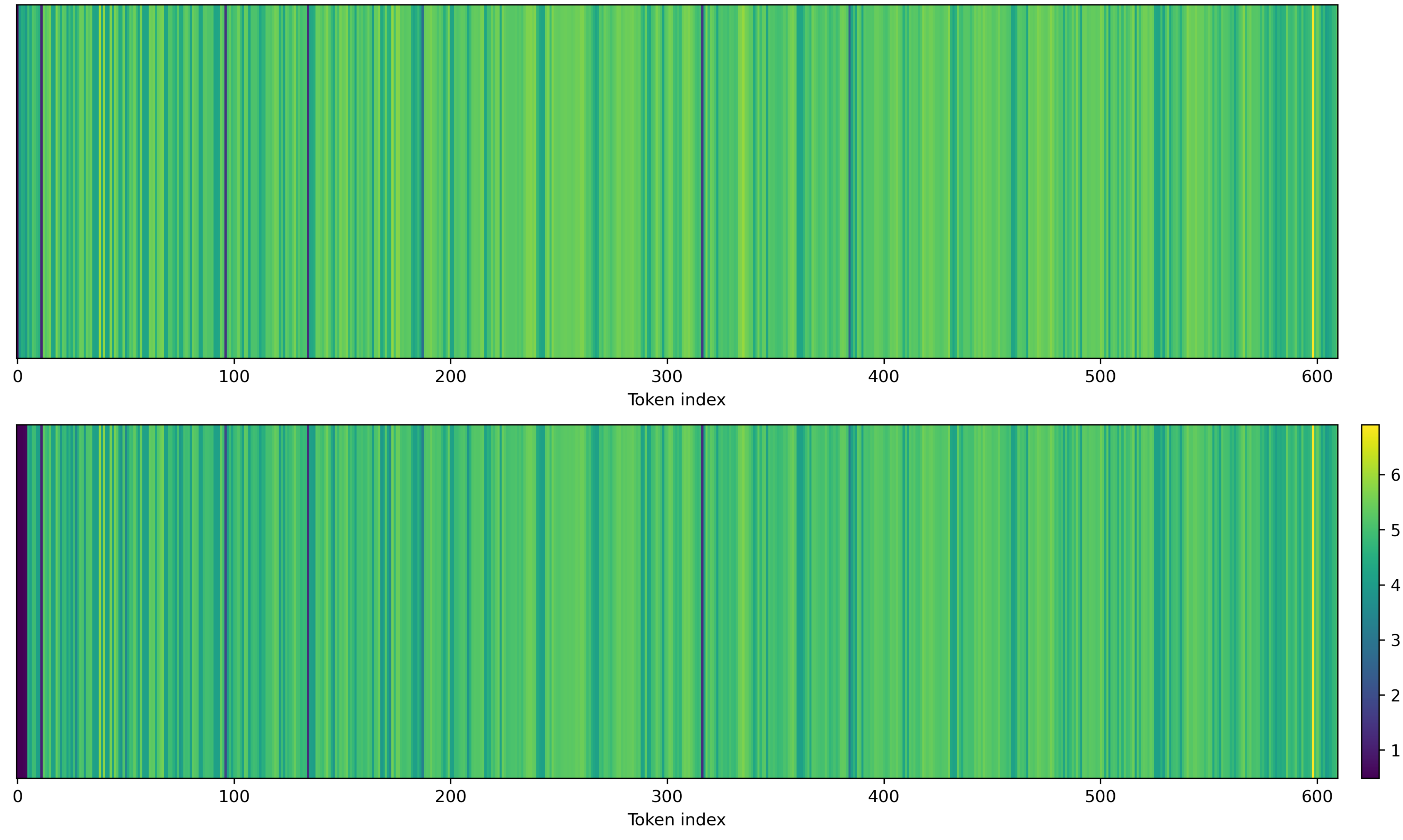}
    \caption{Value of retain $V_\phi$ from the vanilla model (top) and ours (bottom) at the last layer.}
    \label{fig:last_layer_retain}
\end{figure}

\begin{figure}[t]
    \centering
    \includegraphics[width=\linewidth]{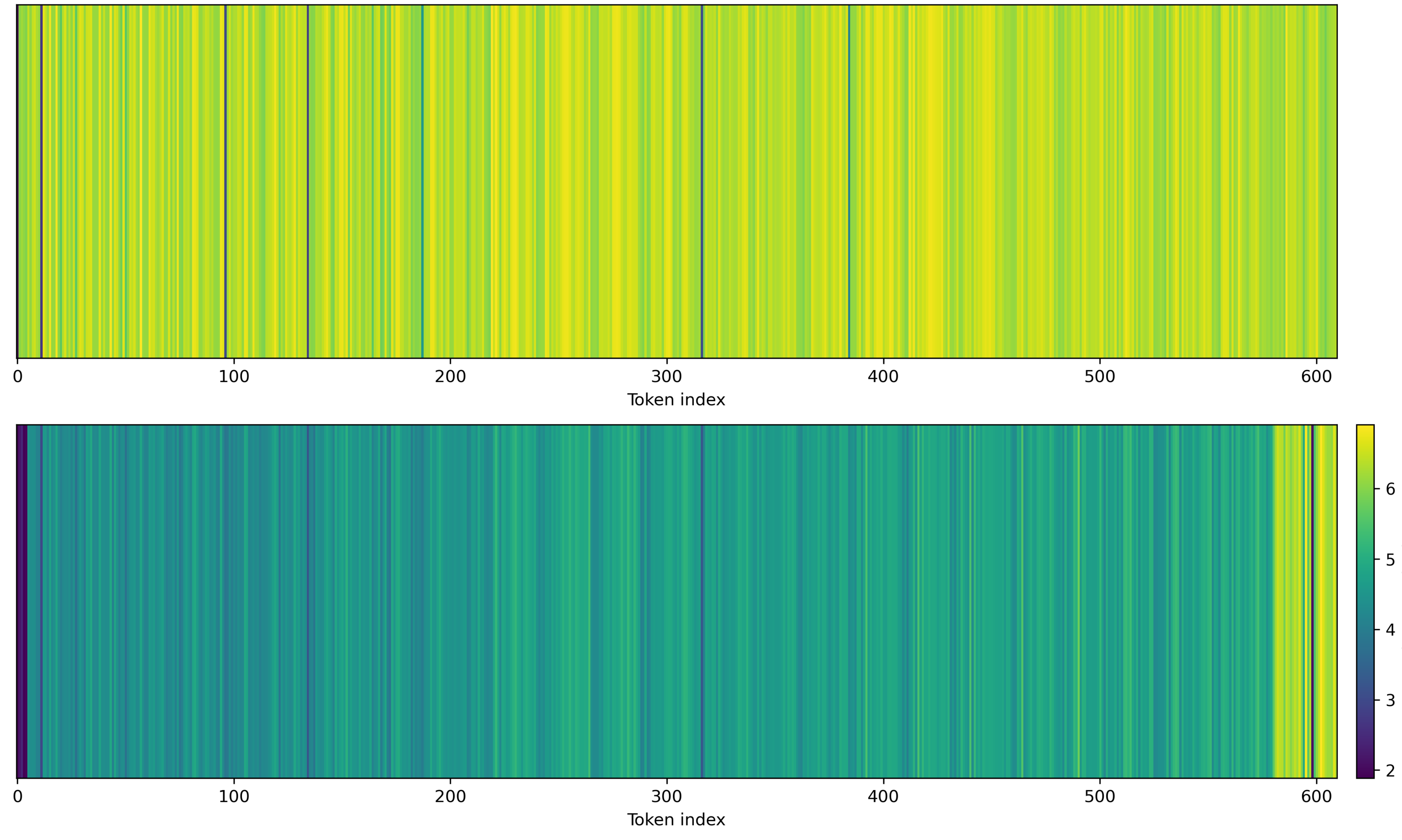}
    \caption{Value of forget $V_\phi$ from the vanilla model (top) and ours (bottom) at the penultimate layer.}
    \label{fig:penultimate_layer_forget}
\end{figure}

\begin{figure}[t]
    \centering
    \includegraphics[width=\linewidth]{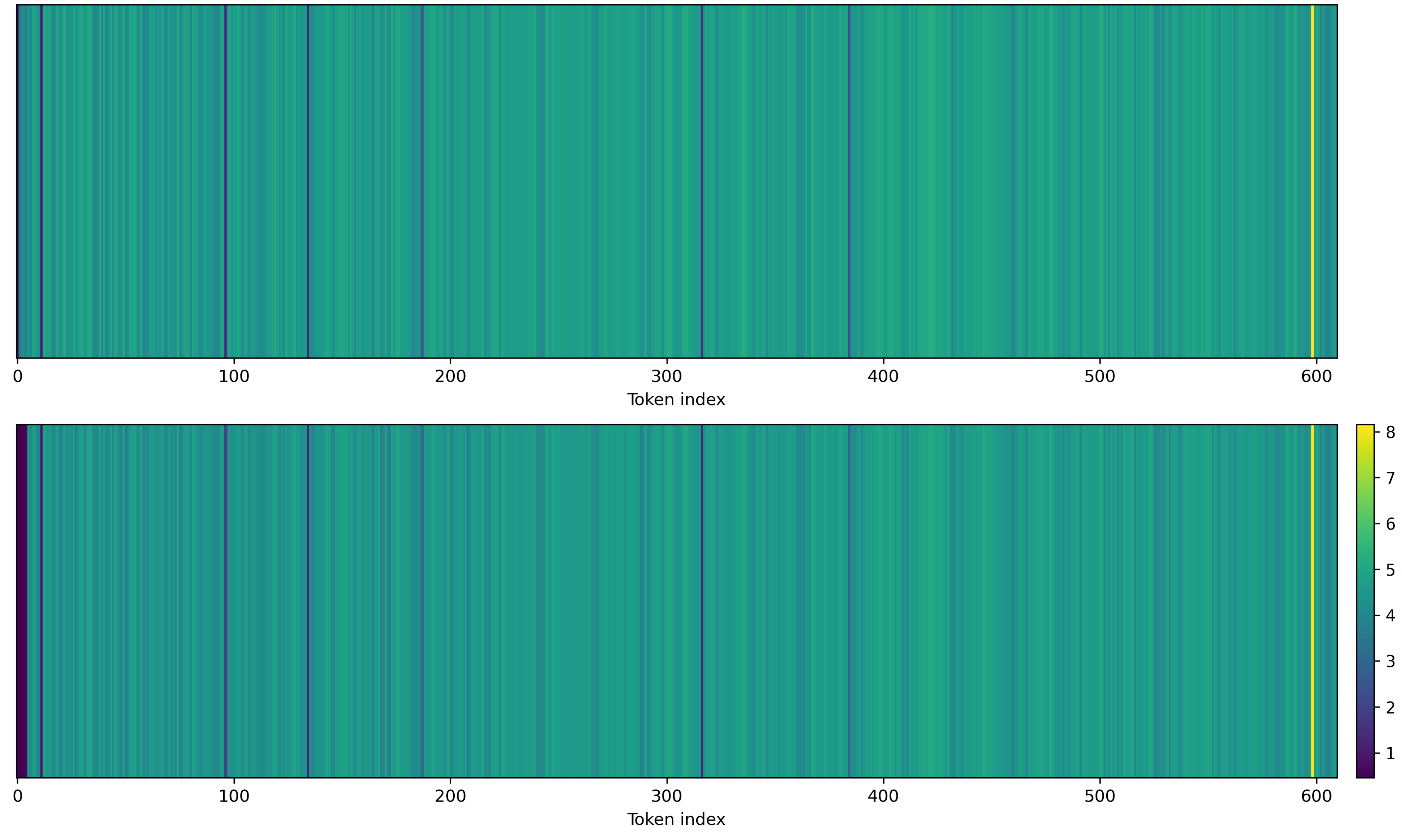}
    \caption{Value of retain $V_\phi$ from the vanilla model (top) and ours (bottom) at the penultimate layer.}
    \label{fig:penultimate_layer_retain}
\end{figure}

We depict the value of $V_\phi$ at the last and penultimate layers of the vanilla model and our model in Figures \ref{fig:last_layer_forget}, \ref{fig:last_layer_retain}, \ref{fig:penultimate_layer_forget}, and \ref{fig:penultimate_layer_retain}. Both models are fine-tuned on top of LLaVA-1.5-7B. The results show that our method substantially suppresses the value activations of forget samples, while the retain samples remain 
comparatively stable. This suggests that the method selectively weakens information carried by the value vectors for forgotten knowledge without broadly disrupting retained representations. Moreover, suppressing the value vectors directly limits the information propagated through attention, making recovery through attention redistribution alone more difficult.





\subsubsection{Ablation Studies on $C_V$}
\label{app:ab_cv}

\begin{figure}[t]
    \centering
    \includegraphics[width=\linewidth]{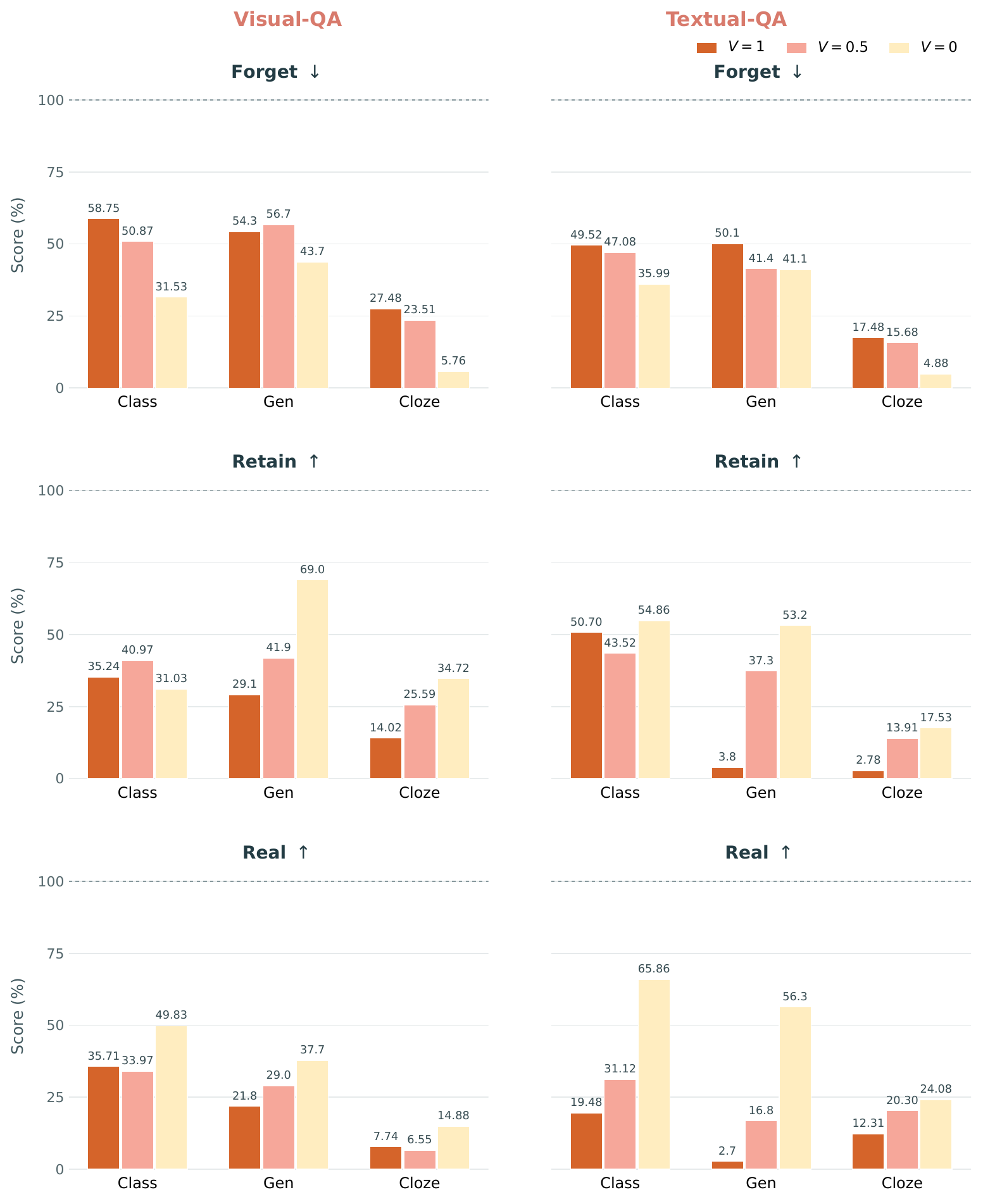}
    \caption{Effect of the attention-value $V$ on LLaVA-1.5-7B. The left and right columns show Visual-QA and Textual-QA results. Each row reports Forget, Retain, and Real performance, with bars for $V\in\{1,0.5,0\}$ grouped by classification accuracy (Class), generation ROUGE-L (Gen), and cloze accuracy (Cloze). ROUGE-L scores are multiplied by 100 for the shared axis. Arrows indicate the preferred performance for evaluation.}
    \label{fig:v_bar}
\end{figure}

Figure \ref{fig:v_bar} shows that reducing \(V\) generally improves forgetting. Forget scores decrease monotonically from \(V=1\) to \(V=0\) in five of the six metrics, and \(V=0\) achieves the lowest forget score in all six. In Visual-QA, forget classification and cloze accuracy fall to 31.53 and 5.76. Generation ROUGE-L also reaches its lowest value, 0.437, although it rises slightly at \(V=0.5\). In Textual-QA, \(V=0\) gives forget scores of 35.99, 0.411, and 4.88 for classification, generation, and cloze. At the same time, \(V=0\) achieves the highest retain score in five of six metrics and the highest Real score in all six. The exception is Visual-QA retain classification, which falls to 31.03. These results make \(V=0\) the strongest observed forget–retain trade-off in this ablation.

\subsubsection{Effect on attention-map}
\label{app:effect_attention_map}

\begin{figure}[t]
    \centering
    \includegraphics[width=\linewidth]{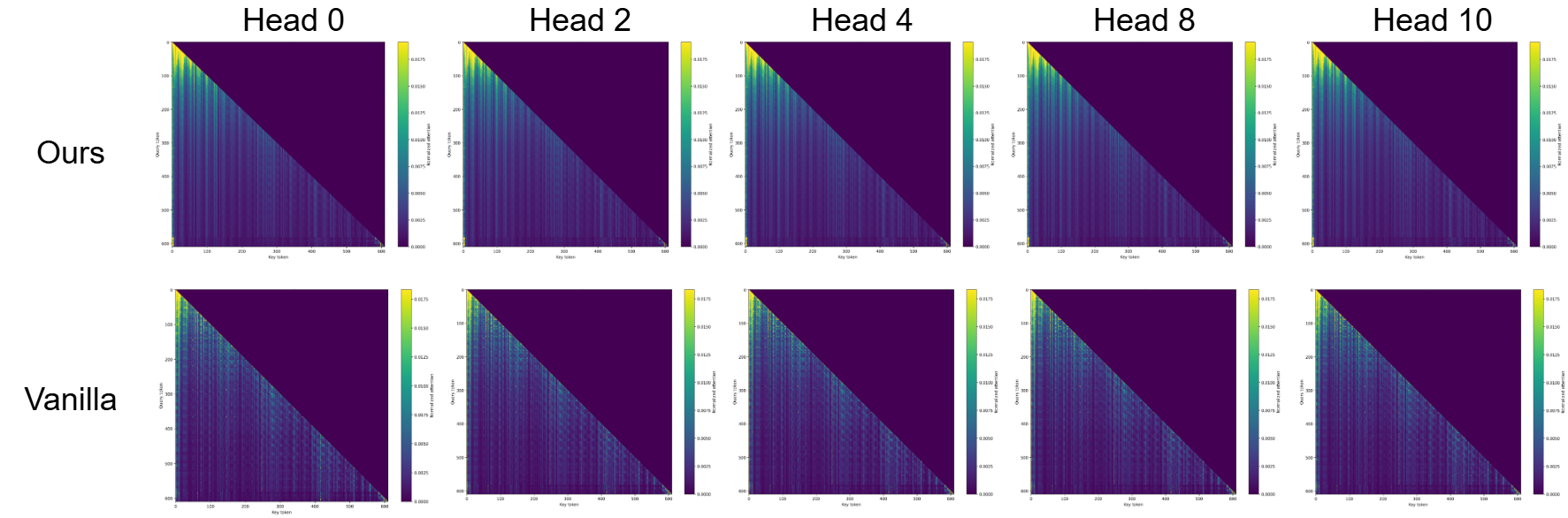}
    \caption{Value of $P_\phi$ from the vanilla model and ours at the last layer.}
    \label{fig:last_layer_p}
\end{figure}

\begin{figure}[t]
    \centering
    \includegraphics[width=\linewidth]{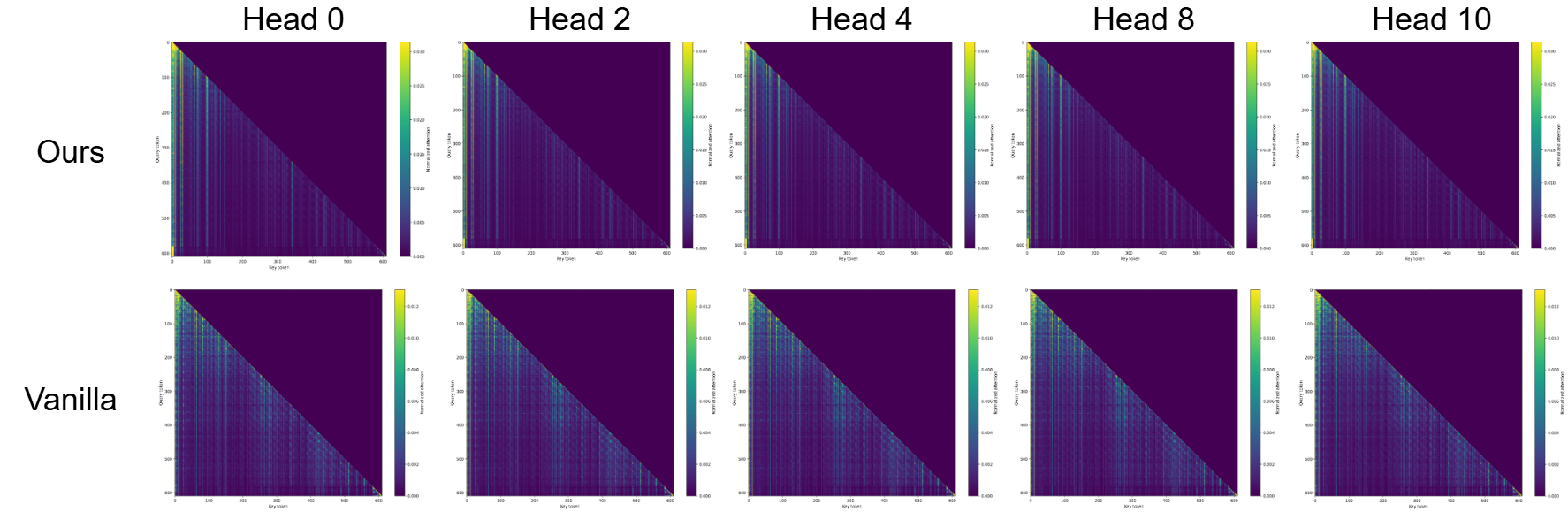}
    \caption{Value of $P_\phi$ from the vanilla model and ours at the penultimate layer.}
    \label{fig:penultimate_layer_p}
\end{figure}

We depict the value of $P_\phi$ at the last and penultimate layers of the vanilla model and our model in Figures~\ref{fig:last_layer_p}, \ref{fig:penultimate_layer_p}. Both models are fine-tuned on top of LLaVA-1.5-7B. Figures~\ref{fig:last_layer_p} and \ref{fig:penultimate_layer_p} compare the attention probability matrix $P_{\phi}$ of our model with the vanilla model at the last and penultimate transformer layers. Across different attention heads, our method largely preserves the original causal attention structure, with highly similar distributions to the vanilla model. Only minor changes in attention intensity are observed within the valid lower-triangular region. This suggests that the proposed unlearning procedure does not substantially disrupt the model's attention routing mechanism. Instead, the model retains a similar pattern of token-to-token interactions after unlearning, indicating that the modification is localized rather than causing a global distortion of the attention distribution.

\section{More related works}
\textbf{Unlearning in language models.}
Machine unlearning aims to reduce the influence of designated training information while preserving the model's remaining capabilities \citep{yao2024largelanguagemodelunlearning,liu-etal-2024-towards-safer,Ren2025SoKMU,pawelczyk2024incontext,yao-etal-2024-machine}. In language models, TOFU introduced a benchmark based on fictitious author profiles and showed that reduced recall of forget data alone does not establish successful unlearning~\citep{maini2024tofu}. A common approach is gradient ascent, which increases the loss on forget examples to discourage their reconstruction. However, unrestricted ascent can lead to substantial degradation of retained behavior. Negative Preference Optimization (NPO) addresses this issue through a reference-relative objective that limits the instability of direct gradient ascent~\citep{zhang2024npo}. Representation Misdirection for Unlearning (RMU) instead acts on internal activations by steering forget representations toward a fixed direction while matching retain representations to those of a frozen reference model~\citep{li2024wmdp}. These works show that effective unlearning requires balancing target suppression with preservation of non-target behavior. SIEVE follows this general objective but focuses on selective multimodal associations and intervenes directly on attention-value representations.

\textbf{Model editing in machine unlearning.} A growing line of work modifies internal representations rather than relying only on output-level objectives \citep{hossain2025investigatingmodeleditingunlearning,luo2026knowledgesmith,guo2025mechanistic,jung-etal-2025-come}. RMU provides an important example by steering forget activations away from their original representations while anchoring retained activations to a frozen reference model~\citep{li2024wmdp}. Model-editing methods provide a related perspective. ROME identifies factual associations in feed-forward modules and performs rank-one parameter updates to modify specific knowledge~\citep{meng2022rome}, while MIP-Editor extends internal path localization to multimodal unlearning~\citep{li2026mipeditor}. These approaches demonstrate that targeted changes to intermediate representations or parameters can alter model behavior without modifying the entire network uniformly.

\end{document}